\documentclass[9pt,twocolumn]{article}
\usepackage[left=.75in,right=.75in]{geometry}
\usepackage{booktabs}
\usepackage{dsfont}
\usepackage{amssymb}

\usepackage{multirow}
\usepackage{tikz}
\usepackage{amsmath}
\usepackage{xurl}
\usepackage{threeparttable}
\usepackage{multirow}
\usepackage{placeins}
\usepackage{subcaption}

\usepackage{xcolor}
\usepackage[linesnumbered,ruled,vlined]{algorithm2e}
\usepackage[all]{xy}
\usepackage{hyphenat}
\usepackage{tablefootnote}
\usepackage{titling}
\usepackage{balance}
\usepackage{hyperref}

\usepackage[
	n,
	operators,
	advantage,
	sets,
	adversary,
	landau,
	probability,
	notions,	
	logic,
	ff,
	mm,
	primitives,
	events,
	complexity,
	asymptotics,
	keys]{cryptocode}

\pretitle{\centering\Large\bfseries}
\posttitle{\par\vspace{1em}}

\preauthor{\centering\normalsize\lineskip 0.5em}
\postauthor{\par}

\predate{\centering\small}
\postdate{\par\vspace{1.5em}}

\begin{document} 

\title{The Mosaic Memory of Large Language Models}

\date{}

\author{
Igor Shilov\footnotemark[1] \quad 
Matthieu Meeus\footnotemark[1] \quad 
Yves-Alexandre de Montjoye
\\[1ex]
\textit{Imperial College London}
}

\maketitle

\def\thefootnote{*}\footnotetext{Equal contribution}\def\thefootnote{\arabic{footnote}}

\begin{abstract}
As Large Language Models (LLMs) become widely adopted, understanding how they learn from, and memorize, training data becomes crucial. Memorization in LLMs is widely assumed to only occur as a result of sequences being repeated in the training data. Instead, we show that LLMs memorize by assembling information from similar sequences, a phenomena we call \emph{mosaic memory}. We show major LLMs to exhibit mosaic memory, with fuzzy duplicates contributing to memorization as much as 0.8 of an exact duplicate and even heavily modified sequences contributing substantially to memorization. Despite models display reasoning capabilities, we somewhat surprisingly show memorization to be predominantly syntactic rather than semantic. We finally show fuzzy duplicates to be ubiquitous in real-world data, untouched by deduplication techniques. Taken together, our results challenge widely held beliefs and show memorization to be a more complex, mosaic process, with real-world implications for privacy, confidentiality, model utility and evaluation.
\end{abstract}

\section{Introduction}

Large Language Models (LLMs) are increasingly transforming various aspects of daily life. They drive advancements across industries by automating routine tasks through applications like virtual assistants and customer service bots~\cite{geramifard2022project,subagja2023improving}; extracting valuable insights from unstructured data~\cite{wadhwa2023revisiting}; generating code to support software development~\cite{austin2021program,chen2021evaluating}; and reinventing information retrieval through conversational AI such as ChatGPT~\cite{openai2022chatgpt}.

LLMs develop their capabilities by absorbing patterns from massive text datasets, allowing them to both generalize concepts and selectively memorize information. Generalization allows LLMs to reason~\cite{wei2022chain} and apply their knowledge to unseen scenarios, while memorization helps them retain factual information. The interplay between generalization and memorization in LLMs has been widely studied from a model utility perspective, examining how memorization contributes to generalization~\cite{khandelwal2019generalization,feldman2020does,tay2022transformer}, facilitates the encoding of factual knowledge~\cite{kandpal2023large,petroni2019language}, and how different data mixtures influence model performance~\cite{xie2024doremi}. 

However, memorization can also be \emph{unintended} and raise concerns. Indeed, LLMs have been shown to memorize personal or confidential information~\cite{carlini2019secret,carlini2021extracting,brown2022does,mireshghallah2022empirical,lukas2023analyzing,mireshghallah2023can}, copyright-protected content~\cite{karamolegkou2023copyright,duarte2024cop,meeuscopyright}, as well as portions of evaluation benchmarks, potentially inflating performance estimates and undermining fair assessment~\cite{oren2023proving,roberts2023cutoff,mirzadehgsm,magar2022data,deng2024investigating,li2024task,dong2024generalization}.

The mechanism by which LLMs memorize training data has predominantly been understood as exact, or verbatim. Indeed, current studies of LLM memorization have shown that a piece of content repeated \emph{exactly} several times in the training data has a much higher chance to be memorized~\cite{kandpal2022deduplicating,lee2022deduplicating,carlini2022quantifying,meeuscopyright}. This memorized content could then be extracted by a potential attacker -- either exactly~\cite{carlini2022quantifying} or approximately~\cite{ippolito2022preventing,peng2023near}. This behavior is often highly undesirable, especially if the extractable content contains private information or is copyright-protected. This understanding has, in turn, led model developers to implement data deduplication techniques to improve model utility~\cite{lee2022deduplicating,hernandez2022scaling,allamanis2019adverse,tirumala2024d4,hoffmann2022training,xue2024repeat}, mitigate privacy and confidentiality risks~\cite{kandpal2022deduplicating}, or to decontaminate benchmark data to properly evaluate model capabilities~\cite{brown2020language,weifinetuned,du2022glam,touvron2023llama2,chowdhery2023palm}. However, the presumed verbatim nature of memorization has led to most mitigation strategies to rely on removing exact repetitions of sequences of text.  

\textbf{Contribution.} We here argue that viewing LLM memorization solely through the lens of exact repetitions in the training data is incorrect. Instead, we show LLMs to have a \emph{mosaic memory}, an ability to memorize sequences from text with partially overlapping fragments (\emph{fuzzy duplicates}). 

First, we introduce a framework to study the mosaic memory of LLMs. Building on established techniques for analyzing text memorization~\cite{kandpal2022deduplicating,meeuscopyright}, we evaluate the performance of Membership Inference Attacks (MIAs) on \emph{canaries} -- artificially crafted sequences included as part of the training data~\cite{carlini2019secret}. We inject a \emph{reference canary}, along with its fuzzy duplicates, into the training dataset of a target LLM, and measure how the presence of these fuzzy duplicates impacts MIA performance on the reference canary. We measure the memorization of fuzzy duplicates using \emph{exact duplicate equivalent} $\rho$, the proportion of the memorization impact retained by the fuzzy duplicate compared to an exact one. A value of $\rho=1$ indicates that a fuzzy duplicate contributes to memorization of the reference canary as much as an exact duplicate, while $\rho=0$ would indicate that the fuzzy duplicate has no impact on the memorization. 

We study four major LLMs: Llama-3.2 (Meta), Phi-2 (Microsoft), Gemma-2 (Google), GPT-Neo (EleutherAI), and find all of them to exhibit a significant mosaic memory. For example, fuzzy duplicates with $10\%$ of their tokens replaced contribute to the memorization of the canary between $\rho=0.50$ (Phi-2 and Gemma-2) and $\rho=0.60$ (GPT-Neo and Llama-3.2) of an exact duplicate's impact. Moreover, even heavily modified fuzzy duplicates -- where $50\%$ of tokens are replaced -- still contribute as much as $\rho=0.15-0.19$ of an exact duplicate. Furthermore, we find that mosaic memory is remarkably robust, as the $\rho$ of fuzzy duplicates remains significantly higher than $0$, even when separated by noise insertions or when their order is shuffled.

Second, while memorization is not exact, we find, somewhat surprisingly, that it is still predominantly \emph{syntactic}, rather than \emph{semantic}. That is, it is driven mostly by the model’s retention of specific overlapping tokens, and not the encoding of the shared underlying meaning across fuzzy duplicates. Fuzzy duplicates created by replacing tokens while preserving semantic meaning indeed are only marginally better memorized than those where semantic meaning is not maintained. Likewise, our results suggest that paraphrased sequences are primarily memorized due to token overlap rather than their shared semantic meaning.

Third, we investigate how widespread fuzzy duplicates are in the real-world, large-scale dataset used for LLM pretraining, SlimPajama~\cite{cerebras2023slimpajama}. Although this dataset is already deduplicated on the document level, we find that many sequences still have a substantial number of both exact and fuzzy duplicates. To identify these fuzzy duplicates, we use Levenshtein distance, which measures the minimum number of single-character edits required to transform one sequence into another. Levenshtein distance 10, for instance, represents just 10\% of tokens edited and corresponds to $\rho$ values between $0.6$ and $0.8$. Our analysis reveals that an arbitrary sequence with 1000 exact duplicates also has, on average, 4000 more fuzzy duplicates at Levenshtein distance 10 (representing just 10\% of tokens edited) and over 20,000 at distance 50. This suggests that traditional deduplication methods are likely insufficient for addressing memorization concerns for LLMs.

Taken together, our results reveal a critical gap in our understanding of how LLMs learn from and memorize training data, whether beneficial or harmful. Rather than solely memorizing exact repetitions, LLMs also retain information across fuzzy duplicates, often to a degree comparable to exact duplicates. This mosaic memory phenomenon fundamentally challenges current assumptions underlying deduplication and privacy protection practices. For example, common industry-standard techniques like removing exact matches from benchmarks (e.g., 13-gram overlaps used in GPT-3 evaluations) fail to eliminate fuzzy duplicates, potentially inflating evaluation scores and undermining fair benchmarking. Similarly, deduplication strategies that target exact duplicates (e.g., substring matching of 50 tokens) neglect mosaic memory, allowing memorization of confidential or copyright-protected content through slightly modified sequences. Thus, current practices offer insufficient privacy protection, incomplete benchmark decontamination, and suboptimal data preprocessing, necessitating more nuanced deduplication techniques that account for the mosaic nature of memorization.
 
\section{LLMs have a mosaic memory}

\begin{figure*}[t]
\centering
\includegraphics[width=0.5\linewidth]{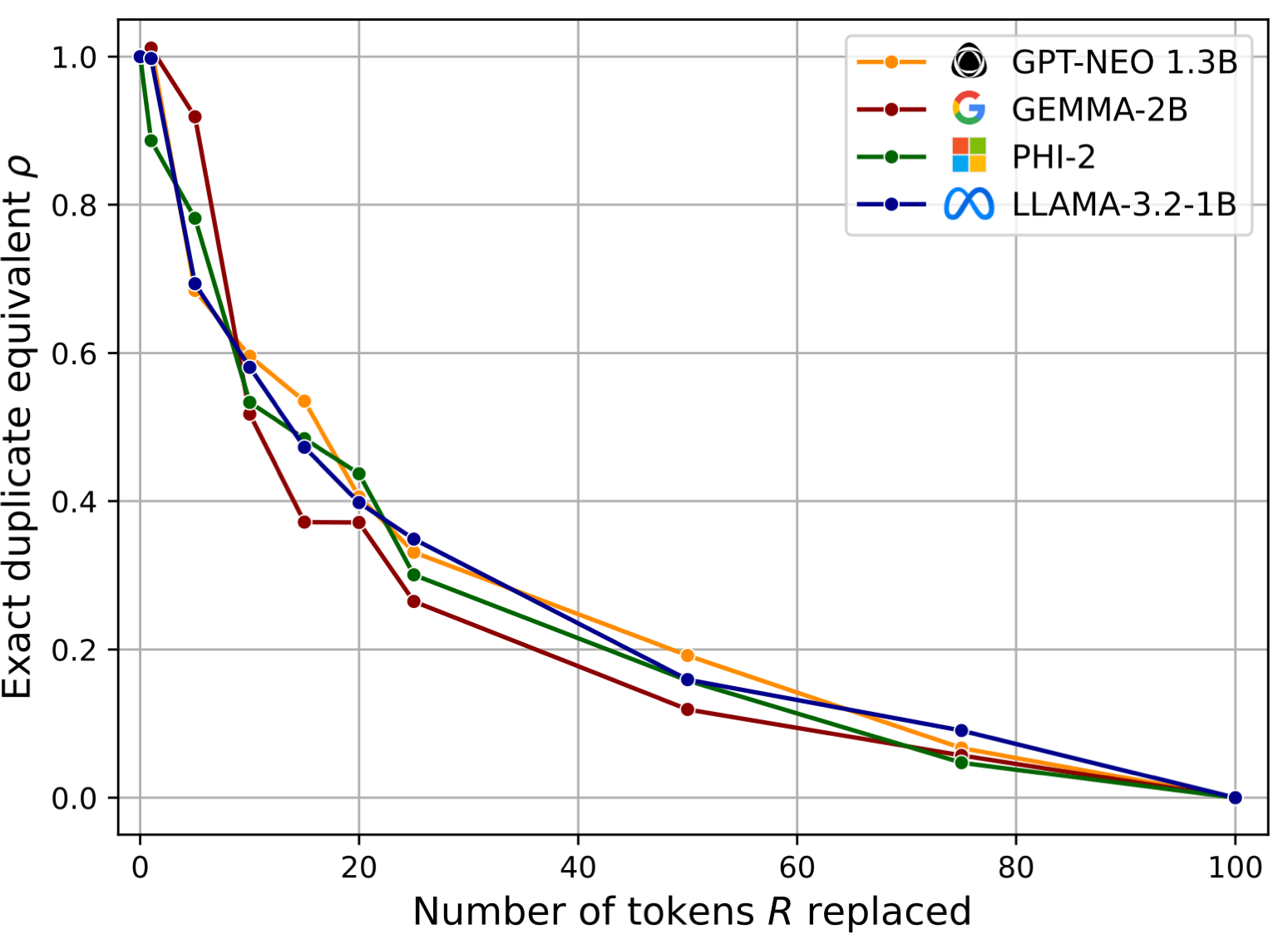} 
    \caption{\textbf{LLMs have a mosaic memory.} The exact duplicate equivalent $\rho$ for fuzzy duplicates across number of replacements made. For smaller values of $R$, fuzzy duplicates contribute to memorization almost equally well than exact duplicates ($\rho=1$) while for larger values of $R$, memorization remains significantly higher than if the canary was entirely absent from the training dataset ($\rho>0$). The mosaic memory is present across widely used models: GPT-NEO 1.3B~\cite{gpt-neo}, Gemma-2B~\cite{team2024gemma}, Phi-2~\cite{javaheripi2023phi} and Llama-3.2-1B~\cite{dubey2024llama}.} 
\label{fig:neq_vs_R_all_models}
\end{figure*} 

\textbf{Quantifying a mosaic memory.} We instantiate a framework to measure how an LLM memorizes fuzzy duplicates, relative to exact duplicates, in its training data.

Following prior work, we quantify memorization of a target language model $\textit{LM}$ with tokenizer $T$ by measuring the performance of MIAs on artificially crafted sequences, known as \emph{canaries}~\cite{carlini2019secret,thakkar2020understanding,thomas2020investigating,zanella2020analyzing,stock2022defending,parikh2022canary,meeuscopyright}. 

We define a set of \emph{reference canaries} $\{X_{\text{ref}}^i \mid i = 1, \ldots, C\}$ where each $X_{\text{ref}}^i$ is a synthetically generated sequence following the approach of \cite{meeuscopyright}, containing exactly $100$ tokens, or $|T(X_{\text{ref}}^i)|=100$. We evaluate their vulnerability to MIAs on the target model $\textit{LM}$. 

To generate \emph{fuzzy duplicates}, we consider any algorithm $\mathcal{A}$ that systematically modifies each reference canary, e.g. by replacing $R$ tokens within the sequence at random. Formally, $\mathcal{A}: X_{\text{ref}}^i \mapsto \{ X_j^i \mid j = 1, \dots, n_{\text{dup}} \}$ produces a set of fuzzy duplicates $X_j^i$ for each $X_{\text{ref}}^i$, where we always consider $X_1^i = X_{\text{ref}}^i$. We explore multiple instantiations of $\mathcal{A}$ and compare their impact on memorization. 

We instantiate an MIA with \emph{member} and \emph{non-member} reference canaries. For each reference canary $X_{\text{ref}}^i$, we flip a fair coin to determine random variable $b_i \sim \{0, 1\}$. If $b_i = 1$, we inject $X_{\text{ref}}^i$ and its corresponding fuzzy duplicates $\{ X_j^i \mid j = 2, \dots, n_{\text{dup}} \}$ into a training dataset $D$; otherwise, neither $X_{\text{ref}}^i$ nor its fuzzy duplicates are included. We then consider a pretrained LLM $\textit{LM}_0$, which we further train on $D$, yielding target model $\textit{LM}$. 

We quantify memorization of a set of fuzzy duplicates via the corresponding MIA performance. We apply MIAs~\cite{yeom2018privacy,carlini2021extracting,shi2023detecting} on $\textit{LM}$, computing a \emph{membership score} $\alpha(X_{\text{ref}}^i)$ for each $X_{\text{ref}}^i$ based on query outputs from $\textit{LM}$. Using these scores and their ground-truth membership labels $b_i$, we compute the receiver operating characteristic area under the curve (ROC AUC), denoted as $\tilde{\phi}$. 

To evaluate how fuzzy duplicates are memorized relative to exact duplicates, we define the equivalent number of exact duplicates $\nu_{\text{eq}}$ as the number of exact repetitions of $X_{\text{ref}}^i$ needed to achieve the same MIA AUC as observed for fuzzy duplicates ($\tilde{\phi}$). This approach aims to be invariant to the absolute level of memorization, which is known to be highly sensitive to factors such as model characteristics (e.g., number of parameters), training procedures (e.g. learning rate), and properties of the sequences (e.g. length, perplexity)~\cite{carlini2022quantifying,meeuscopyright}.

To compute $\nu_{\text{eq}}$, we first compute how exact duplicates are memorized in the same experimental setup. We repeat the same procedure now considering, each time,  $\nu \in \{1, \ldots, n_{\text{dup}}\}$ exact copies of the reference canary $X_{\text{ref}}^i$ injected in training dataset $D_{\nu}$. We measure the resulting MIA performance $\phi_{\nu}$ reached for each value of exact repetitions, and determine $\nu_{\text{eq}}$ as the value of $\nu$ for which $\tilde{\phi} \approx \phi_{\nu{\text{eq}}}$. 

We then compute the \emph{exact duplicate equivalent} $\rho$ for a single fuzzy duplicate by normalizing $\nu_{\text{eq}}$ by the total number of fuzzy duplicates considered. As such, the value of $\rho$ is independent of the number of fuzzy duplicates we consider and reflects the equivalent number of exact duplicates represented by a single fuzzy duplicate generated by $\mathcal{A}$. Additional details are provided in Sections~\ref{sup:ref_canaries}-\ref{sup:computing_rho}. 

\textbf{Fuzzy duplicates with token replacements ($\mathcal{A}_{\text{replace}}$).} Out of the many algorithms  $\mathcal{A}$ which can be used to construct fuzzy duplicates, we first consider $\mathcal{A}_{\text{replace}}$, which constructs fuzzy duplicates of each reference canary $X_{\text{ref}}^i$ by replacing $R$ randomly chosen tokens from $T(X_{\text{ref}}^i)$. For each fuzzy duplicate $X_j^i$, where $j \in \{2, \ldots, n_{\text{dup}}\}$, we define a modification set $\mathcal{R}_j^i \subset \{1, \ldots, |T(X_{\text{ref}}^i)|\}$ of size $|\mathcal{R}_j^i| = R$, consisting of token positions randomly sampled without replacement. For each token position $r \in \mathcal{R}_j^i$, the original token $T(X_{\text{ref}}^i)_r$ is replaced by a different one. For more details we refer to Section~\ref{sup:fuzz_dup}.

Our results show that across a range of widely used, recently developed LLMs -- GPT-Neo~\cite{gpt-neo}, Gemma-2~\cite{team2024gemma}, Phi-2~\cite{javaheripi2023phi}, Llama-3.2~\cite{dubey2024llama} -- fuzzy duplicates contribute significantly to memorization. Figure~\ref{fig:neq_vs_R_all_models} illustrates how $\rho$ varies when the number of replacements $R$ made to the fuzzy duplicates increases. We find the exact duplicate equivalent $\rho$ to decrease gradually with the number of replaced tokens $R$, consistently finding $\rho>0$ until all tokens in the sequence are replaced ($R=100$). For instance, considering GPT-NEO 1.3B, when $R=10$ are replaced (out of the $100$ in total), the fuzzy duplicates still yield exact duplicate equivalents of $\rho=0.60$. In practice this means that having two fuzzy duplicates in the training data with 10\% of the original tokens replaced yields higher memorization in the than one exact repetition. Even for $R=50$ replaced tokens, i.e. half of the original tokens, $\rho$ is still maintained significantly above zero at $0.19$. 

The results are furthermore remarkably consistent across a range of widely used LLMs. While the original training data, benchmark performance and the extent to which the model memorizes likely differs substantially between, for instance, GPT-NEO 1.3B and the more recent Llama-3.2-1B, both LLMs show similar memorization behavior in the presence of fuzzy duplicates. 

In Section~\ref{sup:ablations}, we also show our findings to be consistent  across $3$ state-of-the-art MIA methodologies~\cite{yeom2018privacy,carlini2019secret,shi2023detecting}, different values of initial learning rate, different model sizes and across different strategies to generate reference canaries. We further consider slight modifications to $\mathcal{A}_{\text{replace}}$, considering multiple ways of selecting the tokens to be replaced within the reference canary and across fuzzy duplicates, and find this to have limited impact (Section~\ref{sup:ablations_position}). 

So far, we have exclusively considered fuzzy duplicates obtained by replacing $R$ tokens from the reference canary ($\mathcal{A}_{\text{replace}}$). Using the same framework, we now explore alternative methods for constructing fuzzy duplicates.

\textbf{Fuzzy duplicates with token insertions ($\mathcal{A}_{\text{insert}}$).}  First, we consider inserting random tokens at certain positions in the reference canary. We split the tokenized reference canary $T(X_{\text{ref}}^i)$ into $C_n = \frac{|T(X_{\text{ref}}^i)|}{n}$ equally sized $n$-grams and insert $X_{\text{insert}}$ random tokens between each $n$-gram to generate the fuzzy duplicates. This allows us to assess the robustness of LLM memorization to noise, while preserving the original tokens of the reference canary in the same relative order, across each fuzzy duplicate. 

We compare this approach to the baseline scenario, where $n$-grams are randomly scattered throughout the training dataset ($X_{\text{insert}}=\infty$). This baseline reflects how the memorization of individual, independent subsequences contributes to the memorization of the reference canary. Any additional memorization indicates the model's capacity to ignore inserted random tokens and piece together fragments overlapping across fuzzy duplicates, disregarding the inserted tokens as irrelevant noise.

\begin{figure*}[t]
    \centering
    \subfloat[]{\includegraphics[width=0.4\textwidth]{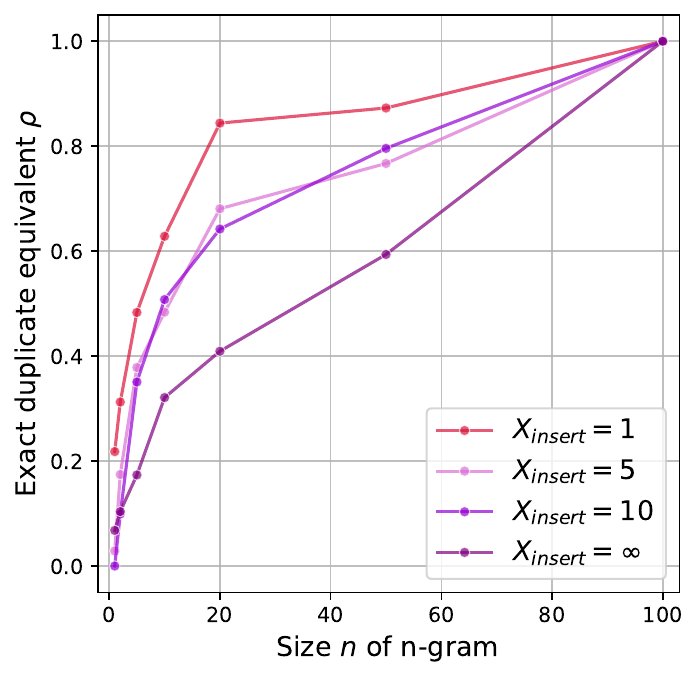}}
    \qquad
    \subfloat[]{\includegraphics[width=0.4\textwidth]{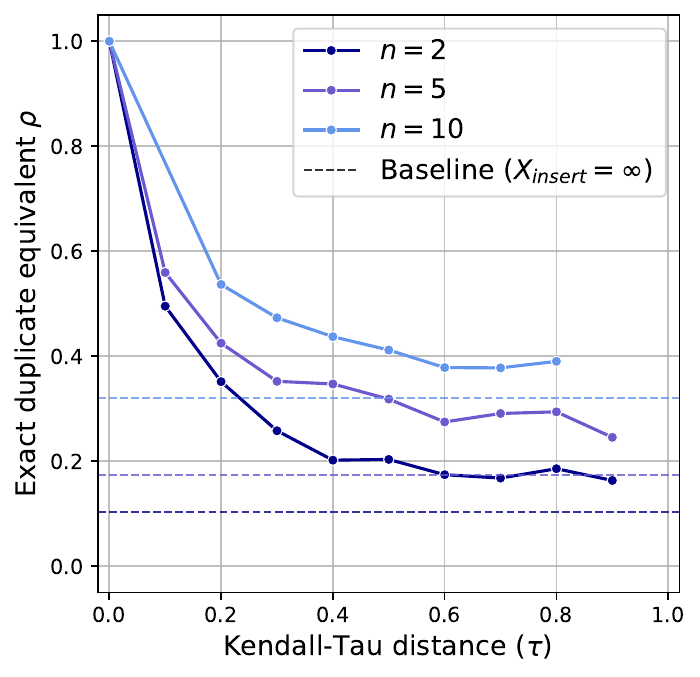}}
    \caption{\textbf{Memorization of fuzzy duplicates constructed with $\mathcal{A}_{\text{insert}}$ and $\mathcal{A}_{\text{shuffle}}$.} (a) The exact duplicate equivalent $\rho$ for fuzzy duplicates when $n$-grams are separated by $X_{\text{insert}}$ tokens ($\mathcal{A}_{\text{insert}}$).  Results demonstrate the model's ability to recognize and memorize content fragments despite the presence of varying amounts of noise tokens inserted between meaningful chunks. Different values of $X_{\text{insert}}$ represent different numbers of random tokens inserted between each $n$-gram, with $X_{\text{insert}} = \infty$ representing the baseline case where $n$-grams are randomly scattered throughout the training dataset. (b) The exact duplicate equivalent $\rho$ for fuzzy duplicates obtained by shuffling $n$-grams ($\mathcal{A}_{\text{shuffle}}$). Results illustrate the impact of token reordering on memorization, with Kendall-Tau distance ($\tau$) measuring the degree of permutation between token pairs. Higher $\tau$ values indicate greater departure from the original sequence order. Different $n$ values represent different sizes of $n$-grams kept intact while their positions were shuffled. Dashed lines show baseline memorization levels for each $n$ value when $n$-grams are randomly scattered ($X_{\text{insert}} = \infty$).}
    \label{fig:unraveling}
\end{figure*}

Figure \ref{fig:unraveling}(a) shows how LLMs exhibit a remarkable robustness in skipping irrelevant tokens. For example, when $X_{\text{insert}}=1$ random token is inserted between all the $20$-grams in the reference canary, the exact duplicate equivalent $\rho$ still reaches $\rho=0.84$, which is significantly larger than the lower bound $\rho=0.41$ at $X_{\text{insert}}=\infty$. When $X_{\text{insert}}=10$ tokens are inserted, the memorization decreases, yet remains as high as $\rho=0.64$. Even in the extreme case where the entire sequence is split into individual tokens ($n=1$), all separated by $X_{\text{insert}}=1$ token, we find a $\rho$ of $0.22$, far exceeding the baseline assumption that such fuzzy duplicates do not contribute to the memorization of the reference canary ($\rho=0$). 

While this remains to be shown, we hypothesize that such robustness derives from the attention mechanism, which could potentially allow the LLM to memorize a connection between tokens even if irrelevant tokens are inserted in between. During training, the LLM would learn to assign low attention scores to inserted tokens, effectively filtering them out as noise. This would lead the model to focus on meaningful patterns across fuzzy duplicates, reinforcing the connections between the original subsequences. When the reference canary’s subsequences then reappear together during inference for the MIA, the model reconstructs these relationships, leading to stronger memorization.

\textbf{Fuzzy duplicates through token shuffling ($\mathcal{A}_{\text{shuffle}}$).} Second, we study the memorization of sequences for which the exact word order is no longer preserved. We now obtain the fuzzy duplicates by shuffling the tokens in the reference canary $X_{\text{ref}}^i$ by varying degrees. 

$\mathcal{A}_{\text{shuffle}}$ first partitions a reference canary $T(X_{\text{ref}}^i)$ into $C = \frac{|T(X_{\text{ref}}^i)|}{n}$ equally sized $n$-grams for $n=\{2,5,10\}$. The fuzzy duplicates are then generated by randomly permuting these $n$-grams while maintaining the original token order within each $n$-gram. From our findings above for $X_{\text{insert}}=\infty$, we observe that individual $n$-grams from the reference canary spread across the dataset contribute non-trivially to memorization. $\mathcal{A}_{\text{shuffle}}$ controls for this effect as it maintains the local order of tokens and shuffle $n$-grams rather than individual tokens. 

To quantify the degree of permutation, we use the token-level \textit{normalized Kendall tau distance} ($\tau$)~\cite{kendall1938new}. Kendall tau measures the proportion of token pairs that maintain their relative order in the fuzzy duplicate compared to the reference canary. Given two sequences of tokens, $T(X_{\text{ref}}^i)$ (reference canary) and $T(X_j^i)$ (fuzzy duplicate obtained by shuffling), Kendall tau is defined as $\tau =\frac{\Delta}{L(L-1)/2}$, where $L=|T(X_{\text{ref}}^i)|$ is the total number of tokens and $\Delta$ is the number of \textit{discordant pairs}, i.e., token pairs $(t_u, t_v)$ where the relative order is different between $X_{\text{ref}}^i$ and $X_j^i$. A value of $\tau=0$ indicates no shuffling (the original order is preserved), while $\tau=1$ implies indicates exact reversal. 

For each $n$, we randomly permute $n$-grams to generate fuzzy duplicates and group them based on their Kendall tau distance. Note that since shuffling $n$-grams instead of individual tokens preserves the local order of tokens, not all $\tau$ values within $[0,1]$ can be reached for every $n$.

Figure~\ref{fig:unraveling}(b) reveals two insights about memorization and token order: memorization is highly sensitive to ordering, yet some information is still memorized even with extensive shuffling. 

First, we observe $\rho$ dropping from $1$ to approximately $0.55$ when only $10\%$ of token pairs have their relative order inverted ($\tau = 0.1$), showing that perturbing the sequence even slightly has a significant impact in memorization.

Second, and more interestingly, we find that the memorization, across all values $\tau$ remains significantly higher than the baseline memorization achieved when $n$-grams are randomly spread across the training dataset ($X_{\text{insert}}=\infty$ from Figure~\ref{fig:unraveling}(a)). Considering 10-grams, for example, we still recover $\rho=0.411$ (against the baseline baseline of $\rho=0.321$) even when the order for $50\%$ of the token pairs is inverted ($\tau=0.5$). These results indicate that the model's mosaic memory retains the shared vocabulary across fuzzy duplicates despite minimal preserved ordering. 

\section{The mosaic memory is syntactic rather than semantic}

All our experiments thus far have focused exclusively on syntactic modifications to create fuzzy duplicates -- systematically altering tokens through replacements, insertions, and shuffling while tracking their impact on memorization. We now examine whether semantic relationships also influence memorization. Our key question is whether LLMs primarily memorize specific tokens or capture underlying meanings. To explore this, we conduct two experiments. First, we systematically vary semantic coherence while controlling token-level changes. Second, we analyze paraphrases as fuzzy duplicates, where semantic meaning is preserved without explicit control over token overlap.

\textbf{Semantic coherence when making $R$ replacements.} First, we consider the same algorithm to make fuzzy duplicates as above ($\mathcal{A}_{\text{replace}}$), where $R$ tokens are replaced for each fuzzy duplicate $X_j^i$ but now vary the extent to which the replacement maintains the semantic meaning of the sequence. In particular, we leverage a masked language model $\textit{MLM}$ to predict the top-$k$ tokens (from the $\textit{MLM}$'s vocabulary $\mathcal{V}_{\textit{MLM}}$) for each replacement and vary the value of $k$, thereby changing how semantically meaningful the replacement is. We thus consider algorithm $\mathcal{A}_{\text{replace}, k}$ where smaller values of $k$ lead to more semantic similarity between the reference canary and the fuzzy duplicates. We use RoBERTa~\cite{liu2019roberta} as $\textit{MLM}$, which has a vocabulary size of $|\mathcal{V}_{\textit{MLM}}|=50,000$. For implementation details and examples of token replacements for varying value of $k$, we refer to Section~\ref{sup:fuzz_dup}.

\begin{figure*}[t]
\centering
\includegraphics[width=0.4\linewidth]{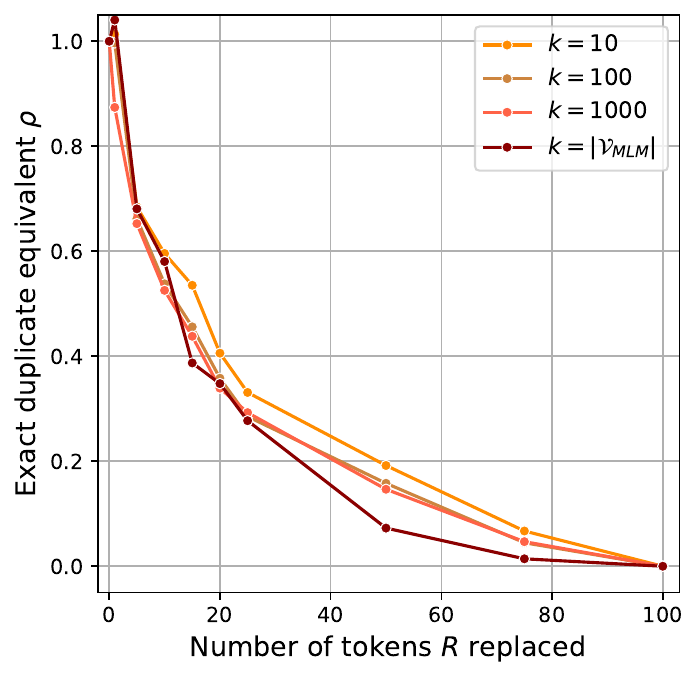} 
    \caption{\textbf{Mosaic memory for varying level of semantic coherence across fuzzy duplicates.} The exact duplicate equivalent $\rho$ for fuzzy duplicates when tokens are replaced with a token sampled from the top $k$ predictions returned by the masked language model $\textit{MLM}$. When $k=10$, tokens are replaced by one of the most likely $10$ tokens predicted by $\textit{MLM}$, while for $k=\mathcal{V}_{\textit{MLM}}$, tokens are effectively replaced by a random token from the MLM's vocabulary.} 
\label{fig:semantic_results}
\end{figure*} 

Figure~\ref{fig:semantic_results} shows the impact of making $R$ replacements with semantically meaningful tokens (smaller $k$) rather than random ones ($k=|\mathcal{V}_{\textit{MLM}}|$). While lower values of $k$ consistently lead to higher memorization, the impact is notably small compared to the impact of syntactic changes.

For example, when $R=20$ tokens are replaced in the fuzzy duplicates, using semantically similar tokens ($k=10$) only increases $\rho$ by $0.06$ ($0.35 \to 0.41$) compared to using a random token ($k=|\mathcal{V}_{\textit{MLM}}|$). This effect size is smaller than replacing just $5$ more tokens while keeping the semantic meaning intact (for $R=25$ and $k=10$, we get $\rho=0.33$).

These results suggest that the mosaic memory of LLMs is more \emph{syntactic} -- the model memorizes the connection between specific, overlapping tokens across the fuzzy duplicates -- than it is \emph{semantic} -- the model memorizes the underlying semantic meaning shared by all fuzzy duplicates. 

\textbf{Comparison to paraphrasing.} We now investigate the memorization of fuzzy duplicates designed to maintain semantic coherence, while not explicitly controlling the token overlap. Indeed, algorithm $\mathcal{A}_{\text{replace}, k}$ is constrained by only replacing a fixed set of tokens, which might not give sufficient flexibility to maintain semantic coherence using just $\textit{MLM}$ replacements. Hence, we now study $\mathcal{A}_{\text{paraphrase}}$, which asks an instruction-tuned LLM to \emph{paraphrase} the reference canary to construct fuzzy duplicates. We consider widely used instruction-tuned LLMs: Llama-3-8B~\cite{llama3modelcard}, Mistral-7B~\cite{jiang2023mistral}, GPT-4o~\cite{openai2024gpt4o}, with additional details provided in Section~\ref{sup:paraphrases}.

\begin{table*}[t]
    \centering
    \begin{tabular}{cccccc}
    \toprule
        & & & \multicolumn{3}{c}{$n$-gram overlap} \\
        \multicolumn{2}{c}{Fuzzy duplicates} & $\rho$ & $n=1$ & $n=2$ & $n=4$ \\
        \hline
        \midrule
        \multirow{3.5}{*}{\parbox{2.5cm}{\centering Paraphrased using}} & Llama-3-8B~\cite{llama3modelcard} & $0.11$ & $39.02\pm19.97$ & $17.68\pm15.39$ & $7.97\pm9.95$\\ 
        \cmidrule{2-6}
        & Mistral-7B~\cite{jiang2023mistral} & $0.17$ & $49.52\pm20.50$ & $26.28\pm17.83$ & $12.85\pm13.87$\\ 
        \cmidrule{2-6}
        & GPT-4o~\cite{openai2024gpt4o}  & $0.30$ & $70.70\pm18.73$ & $45.63\pm23.15$ & $27.89\pm22.89$\\ 
        \midrule
        \bottomrule
    \end{tabular}
    \caption{\textbf{Memorizing paraphrases.} The exact duplicate equivalent $\rho$ for fuzzy duplicates constructed as semantic paraphrases, alongside the mean $n$-gram overlap between the paraphrases and the reference canary. For comparison, we report the results for $n$-gram overlap for fuzzy duplicates with $R$ replacements in Section~\ref{sup:paraphrases}.}
    \label{tab:paraphrases_results}
\end{table*}

Table~\ref{tab:paraphrases_results} shows that for all paraphrases generated using the $3$ instruction-tuned LLMs, the exact duplicate equivalent $\rho$ remains fairly low, ranging from $0.11$ for Llama-3-8B~\cite{llama3modelcard} to $0.30$ for GPT-4o~\cite{openai2024gpt4o}. Moreover, this is remarkably low compared to the memorization achieved when $R$ tokens are replaced with random tokens ($k=|\mathcal{V}_{\textit{MLM}}|$) from Figure~\ref{fig:semantic_results}. For instance, fuzzy duplicates with $20\%$ of all tokens in the reference canary ($R=20$) replaced by random ones -- the semantic meaning is significantly distorted -- still contribute more to the memorization of the reference canary ($\rho=0.35$) than any of the paraphrased fuzzy duplicates. Only when making $R=25$ replacements with random tokens, the level of memorization ($\rho=0.28$) becomes similar to the largest one achieved using paraphrasing ($\rho=0.30$). 

With $\rho$ for paraphrases being already relatively low, we hypothesize that the memorization we observe in this experiment can be explained mostly by syntactic similarity, as paraphrases might still have an overlap in tokens with the reference canary. To investigate this, we compute the mean number of overlapping $n$-grams between each reference canary and its paraphrases for different values of $n$, reporting the results in Table~\ref{tab:paraphrases_results}. We show that, across paraphrases, there remains a substantial $n$-gram (i.e. syntactic) overlap, up to an average of $71$ $1$-grams present in paraphrases made by GPT-4o and that the memorization strictly increases with $n$-gram overlap. Specifically, as the mean overlap in $4$-grams increases from $8.0$ (Llama-3-8B) to $12.9$ (Mistral-7B) and $27.9$ (GPT-4o), the respective $\rho$ also rise from $1.99$ to $2.51$ and $3.72$, respectively -- a trend that holds across different values of $n$. While these results do not rule out that some semantic memorization might happen, taken together they suggest that it likely plays a limited role compared to syntactic memorization.

\section{Fuzzy duplicates in training corpora are widespread and robust to deduplication}

\begin{figure*}[ht]
\centering
\subfloat[]{
\includegraphics[width=0.4\linewidth]{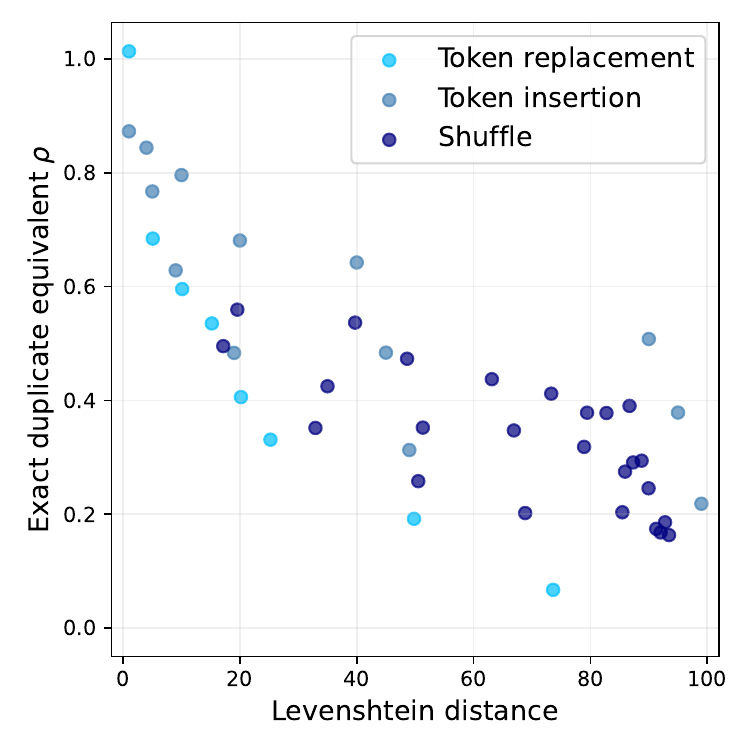}
}
\qquad
\subfloat[]{
\includegraphics[width=0.4\linewidth]{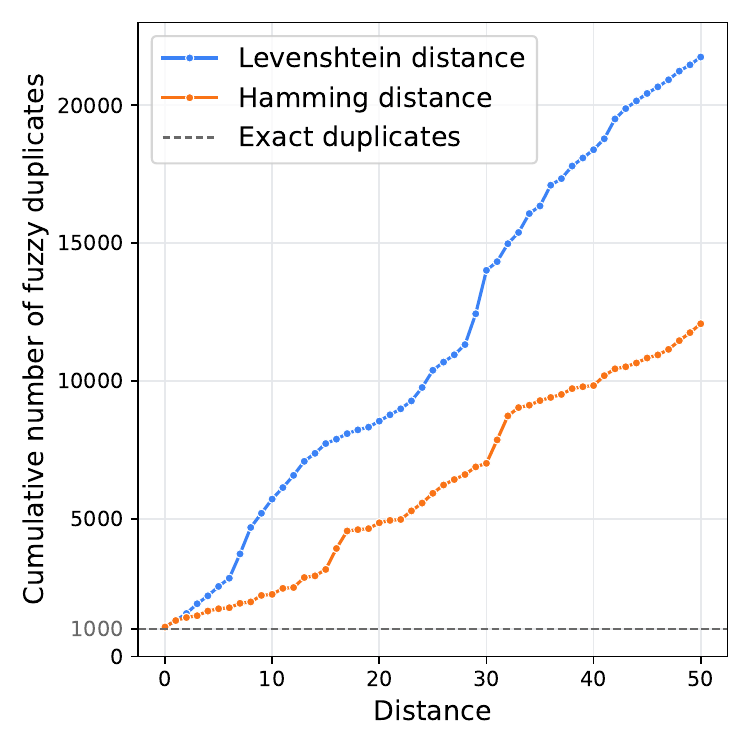}
}
\caption{\textbf{Fuzzy duplicates in SlimPajama.} (a) Correlation between Levenshtein distance and the exact duplicate equivalent $\rho$ based on the experimental results reported for $\mathcal{A}_{\text{replace}}$, $\mathcal{A}_{\text{insert}}$ and $\mathcal{A}_{\text{shuffle}}$. (b) Number of fuzzy duplicates (cumulative) found in SlimPajama with increasing Hamming and Levenshtein distance from the original sequence. Reported numbers are averaged over 100 sequences from a subgroup of sequences repeated verbatim $1,000$ ($\pm 1\%$) times in the dataset.}
\label{fig:slimpajama}
\end{figure*}

Having demonstrated the significant contribution of fuzzy duplicates to model memorization, we now investigate their prevalence in a real-world training dataset.  Our analysis shows that even in the extensively deduplicated datasets a large number of fuzzy duplicates exist, significantly contributing to memorization. 

To identify fuzzy duplicates in real-world datasets, we adopt Levenshtein distance, which measures the minimum number of single-character edits required to transform one sequence into another. It accounts for token replacements ($\mathcal{A}_{\text{replace}}$), insertions ($\mathcal{A}_{\text{insert}}$), and effectively captures token shuffling ($\mathcal{A}_{\text{shuffle}}$) through combinations of deletion and addition operations.

We validate this choice by investigating the relationship between Levenshtein distance and the exact duplicate equivalent ($\rho$) across our experimental results. Figure~\ref{fig:slimpajama}(a) shows a strong negative correlation between $\rho$ and Levenshtein distance, confirming that this metric effectively captures the memorization impact of various fuzzy duplicates.

We also investigate Hamming distance, which counts the number of positions where corresponding tokens differ between two sequences. This metric directly corresponds to $\mathcal{A}_{\text{replace}}$, making fuzzy duplicates identified by Hamming distance real-world examples of our token replacement experiments. 

As a target dataset we consider SlimPajama~\cite{cerebras2023slimpajama} -- a thoroughly cleaned and document-level deduplicated dataset derived from the RedPajama dataset~\cite{together2023redpajama}, containing 627 billion tokens, or 895 GB of text in a compressed form. 

The dataset's size adds additional complexity to the analysis, as computing Levenshtein distance for all possible sequence pairs would be computationally infeasible. Given these constraints, we implement a sampling strategy by identifying a small representative set of sequences and scanning the dataset linearly to find their fuzzy duplicates. For computational reasons we also only scan $5\%$ of the dataset for fuzzy duplicates, and extrapolate the results assuming the uniform distribution of fuzzy duplicate positions -- a reasonable assumption given the dataset is randomly shuffled.

For the target sequence selection we focus on those sequences which are repeated exactly multiple times in the dataset. We do not expect randomly chosen sequences to have many fuzzy duplicates, nor is it necessary to establish our point. Our aim is simply to demonstrate that fuzzy duplicates are sufficiently prevalent to raise memorization concerns. We first confirm that our selection criteria represents a substantial population within the dataset, not just isolated cases. We then analyze these sequences to demonstrate the prevalence of their fuzzy duplicates. If positive, these results would establish the existence of fuzzy duplicates in practice. Note that, however, a thorough quantification of the issue is still required, considering the specifics of the desired objective: privacy and confidentiality protections, benchmark decontamination procedures, or overall model utility considerations.

We hypothesize that the number of fuzzy duplicates correlates with the number of exact repetitions in the dataset. Consequently, we select 100 target sequences that are repeated exactly $1,000$ ($\pm 1\%$) times in SlimPajama and scan for their fuzzy duplicates. We confirm this selection to be meaningful as there are over $700,000$ sequences in the dataset repeated at least $1,000$ times. It makes such sequences sufficiently prevalent so that, if memorized, they would have a major impact on the model's behavior. For implementation details and results from additional repetition frequency buckets ($100$ and $10,000$), we refer to the Section~\ref{sup:slimapajam}. 

Figure~\ref{fig:slimpajama}(b) shows how fuzzy duplicates in SlimPajama increase sharply with Levenshtein distance. Beyond our baseline of $1,000$ exact duplicates (which we explicitly selected for), we find approximately $5,000$ sequences at Levenshtein distance $\leq10$, representing $4,000$ additional fuzzy duplicates. At this distance range, our experiments show exact duplicate equivalent $\rho$ to be between $0.6-1.0$, indicating these fuzzy duplicates significantly contribute to memorization, with their cumulative impact likely to outweigh that of exact repetitions.

As we extend to larger Levenshtein distances up to $50$, corresponding to $\rho$ between $0.2$ and $0.4$, the number of fuzzy duplicates expands to over $20,000$ -- more than 20 times the original count of exact duplicates. Even when using the more conservative Hamming distance metric, which directly corresponds to the token replacements ($R$) in our $\mathcal{A}_{\text{replace}}$ algorithm and provides a direct real-world analog to our controlled canary experiments with predictable memorization effects, we still find approximately $5,000$ fuzzy duplicates at distance $20$ and $10,000$ at distance $40$.


\begin{figure*}[t]
\centering
\includegraphics[width=0.5\linewidth]{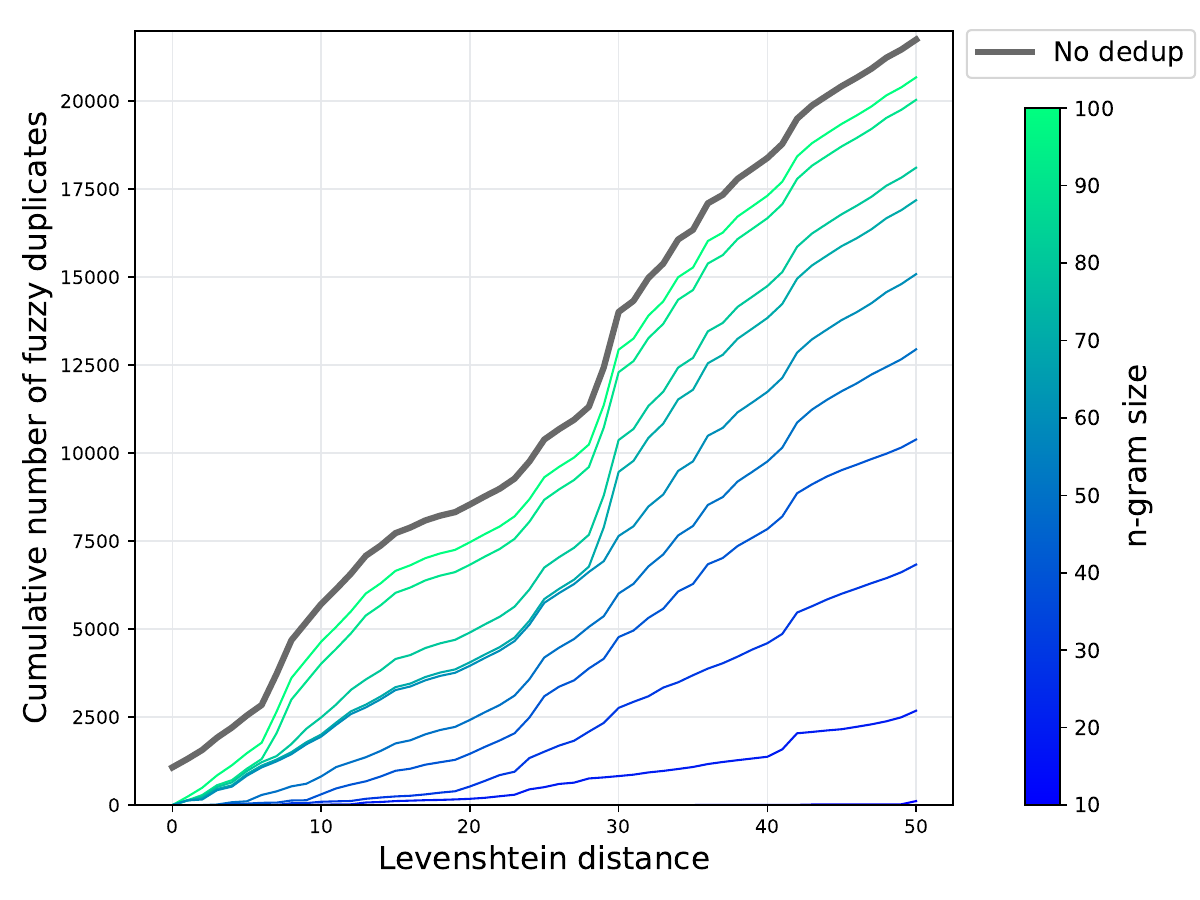} 
    \caption{\textbf{The number of fuzzy duplicates in SlimPajama~\cite{cerebras2023slimpajama} impacted by a varying level of deduplication.} $n$-gram deduplication strategies for varying $n$ fail to account for a range of fuzzy duplicates still contributing substantially to memorization.} 
\label{fig:dedup}
\end{figure*} 

\textbf{Deduplication.} SlimPajama~\cite{cerebras2023slimpajama} has been obtained by deduplicating RedPajama~\cite{together2023redpajama}, removing \emph{documents} with a Jaccard similarity based on $13$-grams higher than $0.8$ (resulting in the removal of 49.6\% of all bytes). Document-level deduplication has been shown to result in a dataset with higher information density~\cite{cerebras2023slimpajama}, allowing developers to train better models with less computational cost~\cite{lee2022deduplicating}. Yet, despite this extensive document-level deduplication, our results showed that SlimPajama still contains a substantial number of exact and fuzzy (sequence-level) duplicates. While training on the same data multiple times -- whether through upsampling or multiple epochs -- can improve model performance, this should be an intentional choice rather than an unintended consequence of duplicated data.

Some model developers have also adopted sequence-level deduplication -- a more expensive, yet more granular alternative to document-level deduplication. Following Lee et al.~\cite{lee2022deduplicating}, a common approach is to remove exactly duplicated substrings of at least $50$ tokens from the training data~\cite{penedo2023refinedweb,kudugunta2024madlad,kandpal2022deduplicating}. Research shows duplicate sequences in training data increase susceptibility to privacy attacks~\cite{carlini2022quantifying,meeuscopyright}, making sequence-level deduplication (typically at 50 tokens) an important mitigation strategy~\cite{kandpal2022deduplicating}. Despite its benefits for dataset efficiency and mitigating privacy risks, sequence-level deduplication requires a more careful calibration than document-level deduplication. Removing sequences may disrupt the internal coherence of documents, and aggressive sequence-level deduplication risks removing a substantial amount of valuable data.

We examine how sequence-level deduplication affects the exact and fuzzy duplicates present in SlimPajama. Specifically, we consider a fuzzy duplicate removed by sequence-level deduplication, if it shares at least one overlapping $n$-gram with the target sequence. While we note that fuzzy duplicates could also be removed due to an overlap with other sequences in the overall dataset, deduplicating the entire dataset comes at a high computational cost. We approximate its effect by considering the set of fuzzy duplicates for increasing Levenshtein distances from Figure~\ref{fig:slimpajama}(b) and argue that the most significant impact is captured by considering this set of highly similar sequences.

Figure~\ref{fig:dedup} shows the number of fuzzy duplicates that remain after sequence-level deduplication for varying values of $n$. We first examine $n=50$ (the fifth curve from the bottom), a common choice in training data preprocessing~\cite{lee2022deduplicating,penedo2023refinedweb,kudugunta2024madlad,kandpal2022deduplicating} or to mitigate privacy risks~\cite{kandpal2022deduplicating}. While deduplicating with $n=50$ removes a substantial amount of fuzzy duplicates -- particularly at low Levenshtein distances ($\leq10$)) --, a significant number remain. 
For instance, at a Levenshtein distance of 20 an average of $2,500$ fuzzy duplicates per sequence persist after deduplication, increasing to $6,000$ for a Levenshtein distance of 30. At the same time, fuzzy duplicates of similar Levenshtein distances still contribute substantially to memorization ($\rho>0.3$ in Figure ~\ref{fig:slimpajama}(a)). These results suggest that standard $50$-gram deduplication likely fails to properly mitigate any privacy risks and might be suboptimal to enhance training efficiency and model performance. While more aggressive sequence-level deduplication (e.g. $n=20$, the bottom curve) might help remove the most impactful fuzzy duplicates, it likely introduces significant trade-offs. 

\section{Discussion}

\textbf{Impact on privacy and confidentiality.} LLMs are trained on extensive corpora of text, typically involving a \emph{pretraining} stage using vast amounts of publicly available internet data, followed by a \emph{posttraining} stage to enhance instruction-following capabilities, reduce harmful outputs, or develop domain-specific expertise~\cite{gpt4techreport,touvron2023llama,touvron2023llama2}. During posttraining, models may be trained on specialized datasets, such as healthcare records~\cite{labrak2024biomistral,liu2023radiology} or human conversations~\cite{ziegler2019fine}, which can introduce personal or sensitive information. Even during pretraining, LLM developers increasingly collect valuable data through licensing deals with, for instance, academic or news publishers~\cite{kwon2024publishers,ftlicense} -- data which may be publicly available but is not necessarily free of sensitive of legally protected information. At the same time, LLMs have been shown to reproduce specific sequences verbatim~\cite{carlini2019secret,carlini2021extracting} and be vulnerable against other privacy attacks such as MIAs~\cite{meeus2024did,shi2023detecting,yeom2018privacy,carlini2019secret}.  

To mitigate privacy risks, training data can be sanitized by removing Personally Identifiable Information (PII) typically using Named Entity Recognition (NER)~\cite{lample2016neural}. NER is however known to be an imperfect process~\cite{vakili2022downstream}, with what constitutes as PII being context-dependent~\cite{brown2022does} and models trained on sanitized data remaining vulnerable to privacy attacks~\cite{lukas2023analyzing,borkar2025privacy}. Regarding confidential information, which may not be directly linked to individuals and thus not picked up by NER, distinguishing what constitutes as confidential information from what benefits model utility is often even more context-dependent, subjective and thus a difficult task. 

Sequence-level deduplication has also been applied as a mitigation strategy~\cite{kandpal2022deduplicating}, as the risk of privacy attacks increases when more exact repetitions of sequences appear in the training data~\cite{carlini2022quantifying,meeuscopyright}. However, our results show that removing all exact occurrences of $50$ tokens might not be sufficient. By design in this work, we study the memorization of fuzzy duplicates by how they contribute to the susceptibility to MIAs, finding that failing to account for them leads to a substantial increase in the risk. As protection against MIAs also implies protection against reconstruction and inference attacks~\cite{salem2023sok}, we reasonably expect our results to generalize to extraction attacks. Furthermore, sequences which are memorized more can not only be extracted verbatim~\cite{carlini2021extracting}, but also in their near-exact form. Recent work indeed argues that studying verbatim regurgitation is too restrictive and does not address the risks associated with the generation of sequences that are highly similar to sequences from the training set~\cite{ippolito2022preventing,peng2023near}.  

\textbf{Deduplication to improve model utility and training efficiency.} Our findings suggest that currently deployed deduplication techniques relying on exact $n$-gram matching for $n=50$~\cite{lee2022deduplicating,penedo2023refinedweb,kudugunta2024madlad,kandpal2022deduplicating} leave many fuzzy duplicates intact. As a result, deduplicated training data might  still contain substantial redundancies, implying suboptimal data preprocessing. More advanced deduplication techniques, for instance based on Levenshtein or Hamming distance, could further improve training efficiency, achieving comparable or superior performance with less data and lower cost. 

Ultimately, the right deduplication technique to reach optimal training efficiency or model utility involves challenging trade-offs that must be carefully considered by model developers in practice. For instance, computing Levenshtein distance across an entire dataset is more expensive than using Hamming distance or exact deduplication. Similarly, determining a deduplication threshold (e.g. value of $n$ in exact deduplication or a maximum distance between sequences) requires balancing data retention with redundancy reduction. 

Further work has also proposed \emph{semantic deduplication}~\cite{abbas2023semdedup} -- removing of fuzzy duplicates based on their similarity in a semantically meaningful embedding space computed using pretrained models. They report efficiency gains up to $15\%$ for smaller language models trained on semantically deduplicated datasets, outperforming exact deduplication~\cite{lee2022deduplicating}. Such semantic deduplication likely comes with similar trade-offs as discussed above. Interestingly, our findings suggest that LLMs tend to memorize more strongly based on syntactic rather than semantic similarity. We hence hypothesize that more aggressive syntactic deduplication strategies may be even more effective than semantic approaches for improving training efficiency. 

\textbf{Benchmark decontamination.} Recent research indicates that the performance reported for newly released LLMs may be inflated and misrepresent progress, as models might be trained on and memorize the benchmarks used to evaluate them~\cite{oren2023proving,roberts2023cutoff,mirzadehgsm,magar2022data,deng2024investigating,li2024task,dong2024generalization,jiang2024investigating}. To mitigate this, sequence-level deduplication techniques based on $n$-gram matching are used for benchmark \emph{decontamination}. Benchmark samples containing $n$-grams overlapping with the training data are removed, and model performance is computed on the remaining samples. This approach was first introduced to evaluate GPT-3~\cite{brown2020language} detecting an overlap of an equivalent of $n=25$ tokens and has since been applied to other models~\cite{weifinetuned,du2022glam}\footnote{These works reportedly use an overlap in $13$ \emph{words} for decontamination, which we here find to correspond to approximately $25$ GPT-NEO tokens.}. This level of deduplication is notably more aggressive than what is commonly used for training data preprocessing ($n=50$).

Figure~\ref{fig:dedup} shows, however, that many fuzzy duplicates contributing to memorization still remain untouched for $n=25$ (the second curve starting from the bottom). Specifically, we recover thousands of fuzzy duplicates at Levenshtein distances of 20–50, where the exact duplicate equivalent remains $\rho \geq 0.2$ (Figure~\ref{fig:slimpajama}(a)). This suggests that the currently deployed decontamination techniques might not be as effective as believed to properly clean benchmarks and ensure the fair evaluation of LLMs, including their reasoning capabilities. 

Subsequent work has explored other techniques for benchmark decontamination. PaLM~\cite{chowdhery2023palm}, for instance, removes samples with at least $70\%$ overlap in $8$-grams, while GPT-4~\cite{gpt4techreport} excludes those for which any of $3$ random sequences of $50$ characters overlap. Llama-2~\cite{touvron2023llama2} filters samples where over $20\%$ of the tokens appear in $10$-grams (at the token level) from the training data, while allowing a `skipgram budget' of $4$ tokens. Finally, a more recent approach suggests determining the decontamination threshold for which model performance diverges the most~\cite{singh2024evaluation}, as used to evaluate Llama-3~\cite{dubey2024llama}. While these methods do not directly map to specific Levenshtein distances, our findings suggest that exact $n$-gram deduplication alone fails to eliminate all fuzzy duplicates. More granular approaches, like those used for Llama-3~\cite{dubey2024llama}, might be more effective, yet have so far not been adopted more widely. 

\textbf{Implications for (adversarial) canaries.} Memorization of canaries has also been used as a tool to infer whether certain content was used to train an LLM, e.g. to detect copyright violations or whether an LLM was trained on an evaluation benchmark. Indeed, it has been proposed to include exact repetitions of unique sequences -- either random token sequences~\cite{wei2024proving} or synthetically generated \emph{copyright traps}~\cite{meeuscopyright} -- , as part of original content. Performing an MIA on these sequences then provides a means of detecting potential copyright violations during model training. Similarly, a unique GUID string was added to the BIG-bench evaluation benchmark~\cite{srivastavabeyond}, both to facilitate model developers to filter out benchmark files from their training data and to enable post-hoc verification of whether an LLM was trained on the benchmark. 

However, exact deduplication would likely remove such canaries. For instance, copyright traps only yield meaningful memorization for 100-token sequences repeated 1,000 times~\cite{meeuscopyright}, which would be trivially removed by commonly deployed deduplication techniques. 

Our findings suggest that the mosaic memory of an LLM could be leveraged to design canaries that are resistant to training data deduplication practices while still being meaningfully memorized by LLMs. Similar techniques could also be leveraged to induce memorization of specific content, including biased opinions or misinformation. 

\bibliography{bibliography.bib} 
\bibliographystyle{plain}

\newpage
\onecolumn

\section*{Acknowledgments}

We thank Marek Rei, Murray Shanahan, Pierre Colombo and Manuel Faysse for useful discussions and feedback.

\section*{Author contributions}
I.S. and M.M. conducted the experiments and analyzed the results. I.S., M.M. and Y.-A. d.M. wrote the paper.

\section*{Data Availability}

All data used in this paper is publicly accessible. To finetune all language models used as target models, we use  books available in the public domain, i.e. the original dataset $D_{\text{orig}}$. We use the open-source library~\cite{kpullygutenberg} to collect $100$ books made available under a permissive license on Project Gutenberg~\cite{projectgutenberg}. We further generate reference canaries by sampling synthetic data from the open-source LLM Llama-2 7B~\cite{touvron2023llama2}. Lastly, we investigate the presence of fuzzy duplicates in real datasets used for LLM training, namely in SlimPajama~\cite{cerebras2023slimpajama}, which is publicly available on the open platform Hugging Face (~\url{https://huggingface.co/datasets/cerebras/SlimPajama-627B}). 

\section*{Code Availability}

The code necessary to reproduce the results in this work has been made publicly available on Github: \url{https://github.com/computationalprivacy/mosaic_memory}. 


\newpage


\renewcommand{\thefigure}{A\arabic{figure}}
\renewcommand{\thetable}{A\arabic{table}}
\renewcommand{\theequation}{A\arabic{equation}}
\renewcommand{\thepage}{A\arabic{page}}
\setcounter{figure}{0}
\setcounter{table}{0}
\setcounter{equation}{0}
\setcounter{page}{1} 
\renewcommand{\thesubsection}{A\arabic{subsection}}


\begin{center}
\section*{Appendix}
\end{center}

\subsection{Reference canaries}
\label{sup:ref_canaries}

We construct a \emph{reference canary} $X_{\text{ref}}^i$ by synthetically generating a sequence using a reference language model $\textit{LM}_{\text{ref}}$ following the approach of Meeus et al.\cite{meeuscopyright}. Starting from an empty string, we iteratively sample the next token from $\textit{LM}_{\text{ref}}$'s predicted probability distribution, using temperature $\mathcal{T}$. We further control for the length of the sequence by truncating the synthetically generated sequence to the first $L_{\text{ref}}$ tokens using the tokenizer $T$ of GPT-NEO~\cite{gpt-neo}, which is the most commonly used target model in this work.  We query $\textit{LM}_{\text{ref}}$ for synthetic sequences repeatedly and apply rejection sampling until we have a sufficient number of reference canaries satisfying $|T(X_{\text{ref}}^i)|=L_{\text{ref}}$. 

As reference model $\textit{LM}_{\text{ref}}$, we use the pretrained Llama-2 7B~\cite{touvron2023llama2} to generate $C=200$ synthetic reference canaries (100 members and 100 non-members), with as length $L_{\text{ref}}=100$. Unless stated otherwise, we use the default temperature $\mathcal{T}=1.0$ to sample from $\textit{LM}_{\text{ref}}$, which we empirically find to lead to meaningful sequences. Only for the results in Figure~\ref{fig:ablation_temperature}, we use other references canaries generated using $\mathcal{T}=2.5$ and $\mathcal{T}=5.0$. Table~\ref{tab:reference_canaries_temp} illustrates some example reference canaries generated with varying sampling temperature.  

\begin{table}[ht]
    \centering
    \begin{tabular}{p{2cm}p{12cm}}
    \toprule
        Temperature & \multicolumn{1}{c}{Selected reference canary $X_{\text{ref}}$} \\
        \midrule
        \midrule
        $\mathcal{T}=1.0$ & \textit{A few years ago I came across a video of the great jazz drummer Art Blakey playing live. The drummer is playing at a ferocious pace yet with an exacting control. I've been practicing my drums more and more, and I wanted to be able to play with that same combination of precision and intensity. I decided to focus on improving my timing by playing a few exercises. The following exercises use quarter notes. The left hand plays with the metronome while the right hand keeps a stead} \\
        \midrule
        $\mathcal{T}=2.5$ & \textit{If not this one then another. At last a new season and a little relief, with some rain! What better time for me in Ireland to be at Bury Street Chapel again for their Winter show; and then what if not with those wonderful, quintessently Australian plants with lots of the lovell colour you expect from plants from 'Oceania Down unda': orchidea in this very pretty mixture and so varied aromathics from grevilleae-rosidiums through the} \\
        \midrule
        $\mathcal{T}=5.0$ & \textit{To get to Tierra Corintiano 5 you got up at o'dumb dark thirty for breakfast, and I took it down on my pants cuphone in the trades to my partner down near Mogotes Cable One—Betina Maribuen (nicknam' Big Butcher). On a line about 6 k a go. She can play music in that place about ¥½ the usual prices; then charge up the 116 wonders why when you get down. We get this in for a quickie then} \\
        \midrule
        \bottomrule
    \end{tabular}
    \caption{Examples of reference canaries, synthetically generated with Llama-2 7B as $\textit{LM}_{\text{ref}}$ while varying the sampling temperature $\mathcal{T}$.}
    \label{tab:reference_canaries_temp}
\end{table}

\subsection{Generating fuzzy duplicates}
\label{sup:fuzz_dup}

To generate fuzzy duplicates, we consider any algorithm $\mathcal{A}: X_{\text{ref}}^i \mapsto \{ X_j^i \mid j = 1, \dots, n_{\text{dup}} \}$ which systematically modifies reference canary $X_{\text{ref}}^i$ to construct a set of fuzzy duplicates $X_j^i$.  We always ensure $X_1^i = X_{\text{ref}}^i$ and use $n_{\text{dup}}=10$. We consider a range of algorithms $\mathcal{A}$ throughout this work, and formalize each of them below. 

\textbf{Replacing $R$ tokens ($\mathcal{A}_\text{replace}$).} Throughout many experiments in this work (e.g. Figure~\ref{fig:neq_vs_R_all_models}), we make $R$ token replacements to construct the fuzzy duplicates from the tokenized reference canary $T(X_{\text{ref}}^i)$, consisting of exactly $L_{\text{ref}}$ tokens. 

For each fuzzy duplicate $X_j^i$, where $j \in \{2, \ldots, n_{\text{dup}}\}$, we define a modification set $\mathcal{R}_j^i \subset \{1, \ldots, |T(X_{\text{ref}}^i)|\}$ of size $|\mathcal{R}_j^i| = R$, consisting of token positions randomly sampled without replacement. To make token replacements, we consider replacing $R = \{1, 5, 10, 15, 20, 25, 50, 75\}$ of the $100$ tokens in the tokenized reference canary.

For each token position $r \in \mathcal{R}_j^i$, the original token $T(X_{\text{ref}}^i)_r$ is replaced by a new token sampled from the top-$k$ most probable tokens predicted by the masked language model $\textit{MLM}$. Specifically, given the masked sequence $T(X_{\text{ref}}^i)_{\setminus r}$, where the token at position $r$ is removed, $\textit{MLM}$ assigns a probability distribution $P_{\textit{MLM}}(\cdot \mid T(X_{\text{ref}}^i)_{\setminus r})$ over the vocabulary. The replacement token is then drawn uniformly at random from the top-$k$ most probable tokens according to $P_{\textit{MLM}}$. Thus, each fuzzy duplicate $X_j^i$ is generated as:

\begin{equation}
    T(X_j^i)_r \sim \text{Top-}k \left( P_{\textit{MLM}}(\cdot \mid T(X_{\text{ref}}^i)_{\setminus r}) \right), \quad \forall r \in \mathcal{R}_j^i.
\end{equation}

Note that, when $k$ equals the size of the entire vocabulary $\mathcal{V}_{\textit{MLM}}$ of tokenizer $T_{\textit{MLM}}$, or $k=|\mathcal{V}_{\textit{MLM}}|$, we replace the token with a randomly sampled token from $\mathcal{V}_{\textit{MLM}}$. This likely distorts the fluency and semantic meaning of the sequence. For smaller values of $k$, the token is replaced by a token which corresponds to a higher predicted probability returned by $\textit{MLM}$, likely maintaining the fluency and semantic meaning of the sequence. We use RoBERTa~\cite{liu2019roberta} as $\textit{MLM}$, which has a vocabulary size of $|\mathcal{V}_{\textit{MLM}}|=50,000$. 

Table~\ref{tab:examples} illustrates one synthetically generated reference canary and two fuzzy duplicates with $R=5$ replacements made for both $k=10$ and $k=|\mathcal{V}_{\textit{MLM}}|$. We find that when $k=10$, i.e. we replace the token by the token with a high predicted probability according to the $\textit{MLM}$, the fluency and semantic meaning of the sequence is well preserved. For instance, replacing \textit{`control'} by \textit{`tempo'} does not alter the fluency of the sequence. In contrast, when $k=|\mathcal{V}_{\textit{MLM}}|$, we are effectively inserting random tokens from the model vocabulary, which quickly distorts the fluency and meaning of the sequence. 

\begin{table}[!ht]
    \centering
    \begin{tabular}{p{0.1cm}p{7cm}p{7cm}}
    \toprule
         & \multicolumn{2}{c}{Token replacement strategy} \\
        $R$ & \multicolumn{1}{c}{$k=10$} & \multicolumn{1}{c}{$k=|\mathcal{V}_{\text{LM}}|$ (random)} \\
        \midrule
        \midrule
        0 & \multicolumn{2}{p{14cm}}{\textit{\underline{\textbf{A}} few years ago I came across a video of the great jazz drummer Art Blakey playing live. The drummer is playing at a ferocious pace yet with an exacting \underline{\textbf{control}}. I've been practicing my drums more and more, and \underline{\textbf{I}} wanted to be able to play with that same combination of precision and intensity. I decided to focus on improving my timing by playing a few exercises\underline{\textbf{. }}The following exercises use quarter notes. The left hand plays \underline{\textbf{with}} the metronome while the right hand keeps a steady}} \\
        \midrule
        \midrule
        5 & \textit{\underline{\textbf{Some}} few years ago I came across a video of the great jazz drummer Art Blakey playing live. The drummer is playing at a ferocious pace yet with an exacting \underline{\textbf{tempo}}. I've been practicing my drums more and more, and \underline{\textbf{immediately}} wanted to be able to play with that same combination of precision and intensity. I decided to focus on improving my timing by playing a few exercises\underline{\textbf{. `}}The following exercises use quarter notes. The left hand plays \underline{\textbf{to}} the metronome while the right hand keeps a steady} & \textit{\underline{\textbf{ington}} few years ago I came across a video of the great jazz drummer Art Blakey playing live. The drummer is playing at a ferocious pace yet with an exacting \underline{\textbf{']}} I've been practicing my drums more and more, and \underline{\textbf{nephew}} wanted to be able to play with that same combination of precision and intensity. I decided to focus on improving my timing by playing a few exercises\underline{\textbf{Facebook }}The following exercises use quarter notes. The left hand plays \underline{\textbf{Marg.}} the metronome while the right hand keeps a steady} \\
        \midrule
        \bottomrule
    \end{tabular}
    \caption{Examples of reference and corresponding fuzzy duplicates for $R=5$ and varying $k$.}
    \label{tab:examples}
\end{table}

Lastly, to generate the fuzzy duplicates using $\mathcal{A}_\text{replace}$, we can additionally vary which tokens from the original reference canary are replaced. Specifically, we define the modification sets $\mathcal{R}_j^i \subset \{1, \ldots, |T(X_{\text{ref}}^i)|\}$, where each $\mathcal{R}_j^i$ consists of $R$ token positions subject to different selection strategies. We distinguish between the following cases:

\begin{enumerate}
    \item \texttt{evenly spread + consistent}: The $R$ replacement positions are chosen to be \textit{evenly distributed} across the reference canary. The sequence is partitioned into $R$ contiguous segments of (approximately) equal size, and one token position is selected at random from each segment. Formally, we define  
    \begin{equation}
    S_i = \{S_{i,1}, S_{i,2}, \dots, S_{i,R}\}, \quad \text{where } S_{i,r} = \left[\frac{(r-1)}{R}|T(X_{\text{ref}}^i)|, \frac{r}{R}|T(X_{\text{ref}}^i)| \right) 
    \end{equation}
    as the $r$-th segment of token indices. The global modification set $\mathcal{R}_i$ is then sampled as $
    \mathcal{R}_i = \{ r_{i,1}, r_{i,2}, \dots, r_{i,R} \}$, where $r_{i,r} \sim S_{i,r}$.
    This set is used consistently across all fuzzy duplicates, i.e., $
    \mathcal{R}_j^i = \mathcal{R}_i, \forall j \in \{2, \ldots, n_{\text{dup}}\}$.

    \item \texttt{not evenly spread + consistent}: The $R$ replacement positions are sampled uniformly at random over token positions, forming one global modification set $\mathcal{R}_i$. This set is then used consistently across all fuzzy duplicates, i.e., $
    \mathcal{R}_j^i = \mathcal{R}_i, \forall j \in \{2, \ldots, n_{\text{dup}}\}$.
    \item \texttt{not evenly spread + not consistent}: The $R$ replacement positions are sampled uniformly at random over token positions, but independently for each fuzzy duplicate. That is, for each duplicate $X_j^i$, we independently sample $\mathcal{R}_j^i$ for each $j \in \{2, \ldots, n_{\text{dup}}\}$. This results in different token replacement sets across fuzzy duplicates.
\end{enumerate}

Throughout this work, unless specified otherwise, we construct fuzzy duplicates using $\mathcal{A}_\text{replace}$ as we consider the most realistic to appear in real-world datasets. For this, we replace tokens with a semantically meaningful other token (opting for $k=10$), not uniformly spread across the reference canary and not consistent across fuzzy duplicates (\texttt{not evenly spread + not consistent}). We study multiple values of $k$ in Figure~\ref{fig:semantic_results} and different ways of replacing tokens across fuzzy duplicates in Figure~\ref{fig:ablations_position}(b). 

\textbf{Inserting random tokens ($\mathcal{A}_\text{insert}$).} For each reference sequence $X_{\text{ref}}^i$, we first partition its tokenized form $T(X_{\text{ref}}^i)$ into $C_n$ contiguous n-grams, where $C_n = \frac{|T(X_{\text{ref}}^i)|}{n}$. To generate fuzzy duplicates, we insert $X_{\text{insert}}$ tokens between each $n$-gram. The inserted tokens are sampled independently from the vocabulary of the target model, or $
I_c \sim \text{Uniform}(\mathcal{V}_{\textit{LM}})^{X_{\text{insert}}}$. The fuzzy duplicate is then constructed as  

\begin{equation}
T(X_j^i) = \bigcup_{c=1}^{C_n-1} \left( T(X_{\text{ref}}^i)_c \cup I_c \right).
\end{equation}

Insertions are resampled for each fuzzy duplicate $X_j^i$, for $j \in \{2, \ldots, n_{\text{dup}}\}$. We study the memorization of fuzzy duplicates constructed using $\mathcal{A}_\text{insert}$ in Figure~\ref{fig:unraveling}(a).

\textbf{Shuffling tokens ($\mathcal{A}_\text{shuffle}$).} For each reference sequence $X_{\text{ref}}^i$, we again partition its tokenized form $T(X_{\text{ref}}^i)$ into $C_n$ contiguous n-grams, where $C_n = \frac{|T(X_{\text{ref}}^i)|}{n}$. To generate fuzzy duplicates with varying degrees of token order permutation, $\mathcal{A}_\text{shuffle}$ randomly swaps these $n$-grams while maintaining the original token order within each $n$-gram.

Formally, let $G = \{g_1, g_2, \ldots, g_{C_n}\}$ represent the sequence of $n$-grams from the tokenized reference canary, where each $g_j$ is an ordered sequence of $n$ consecutive tokens. $\mathcal{A}_\text{shuffle}$ generates a fuzzy duplicate $X^i_j$ by creating a permutation $\pi$ of indices $\{1, 2, \ldots, C_n\}$ and concatenating the $n$-grams in the permuted order: $T(X^i_j) = g_{\pi(1)} \circ g_{\pi(2)} \circ \ldots \circ g_{\pi(C_n)}$.

To quantify the degree of permutation between the reference canary and each fuzzy duplicate, we use the normalized Kendall tau distance $\tau$, which measures the proportion of token pairs whose relative order differs between the two sequences. For token positions $u$ and $v$, let $(t_u, t_v)$ represent a token pair in the reference canary. This pair is considered discordant if its relative order in the fuzzy duplicate is inverted. The Kendall tau distance is defined as:

\begin{equation}
\tau = \frac{\Delta}{L(L-1)/2}
\end{equation}

where $L = |T(X^i_{\text{ref}})|$ is the total number of tokens and $\Delta$ is the number of discordant pairs.

For our experiments, we generate fuzzy duplicates through rejection sampling, continuously permuting $n$-grams and computing $\tau$ until we have obtained the desired number of fuzzy duplicates for a given Kendall tau distance.

\subsection{Membership inference game.} 
\label{sup:mias}

Throughout this work, we measure memorization of a target language model $\textit{LM}$ with tokenizer $T$ by instantiating a membership inference attack (MIA) on the artificially crafted reference canaries. We consider $C=200$ \emph{reference canaries} $\{X_{\text{ref}}^i \mid 1 \ldots, C\}$ with its fuzzy duplicates $\{X_j^i \mid j=2, \ldots, n_{\text{dup}}\}$, always considering $X_1^i$ equal to $X_{\text{ref}}^i$.

We further pretrain $\textit{LM}_0$ on the training dataset $D$, denoting the resulting finetuned model as the target model $\textit{LM}$. For each reference canary $X_{\text{ref}}^i$, we flip a fair coin to determine random variable $b_i \sim \{0, 1\}$. If $b_i = 1$, we inject $X_{\text{ref}}^i$ and its corresponding fuzzy duplicates $\{X_j^i\}_{j=2}^{n_{\text{dup}}}$ into $D$. Conversely, if $b_i = 0$, neither $X_{\text{ref}}^i$ nor its fuzzy duplicates are included in the training dataset. Formally, the training dataset $D$ is defined as:

\begin{equation}
    D = D_{\text{orig}} \cup \left( \bigcup_{i=1}^{C} \left( \{X_j^i\}_{j=1}^{n_{\text{dup}}} \cdot b_i \right) \right),
\end{equation}

where $D_{\text{orig}}$ represents the original dataset without any canary injections, and $\cdot$ denotes set inclusion conditioned on $b_i = 1$.

To quantify memorization, we apply MIAs~\cite{yeom2018privacy,carlini2021extracting,shi2023detecting} on $\textit{LM}$, computing a \emph{membership score} $\alpha(X_{\text{ref}}^i)$ for each $X_{\text{ref}}^i$ based on query outputs from $\textit{LM}$. Each MIA methodology corresponds to a membership scoring function, $\alpha(X_{\text{ref}}^i)$. We select three methods to compute $\alpha(X_{\text{ref}}^i)$: 

\begin{enumerate}
    \item \textbf{\textit{Loss}} attack from~\cite{yeom2018privacy}, which uses the model loss computed on the reference canary. For the sequence of textual characters $X$, tokenized as $T(X) = \{t_1,\ldots,t_L\}$, we denote the loss of language model $\textit{LM}$ with tokenizer $T$ as $\mathcal{L}_{\textit{LM}}(X) = -\frac{1}{L}\sum_{i=1}^{L} \log\left(\textit{LM}_{\theta}(t_i | t_1 \ldots, t_{i-1})\right)$. For the \textbf{\textit{Loss}} attack, we thus consider: $\alpha = \mathcal{L}_{\textit{LM}}(X)$. 
    \item \textbf{\textit{Ratio}} attack from~\cite{carlini2021extracting}, which uses the model loss divided by the loss computed using a reference model, or $\alpha = \mathcal{L}_{\textit{LM}}(X) / \mathcal{L}_{\textit{LM}_{\text{ref}}}(X)$. We use the same $\textit{LM}_{\text{ref}}$ as used to generate reference canaries, i.e. Llama-2 7B~\cite{touvron2023llama2}.
    \item \textbf{\textit{Min-K\% Prob}} from~\cite{shi2023detecting}, which computes the mean log-likelihood of the k\% tokens with minimum predicted probability in the sequence. More formally, $\alpha=\frac{1}{E}\sum_{t_i \in Min-K\%} \log\left( \textit{LM}_{\theta}(t_i)\right)$, where $E$ is the number of tokens in $Min-K\%$ and we consider $k=20$.
\end{enumerate}

The MIAs are evaluated over all $C$ reference canaries, representing a balanced set of \emph{member} canaries ($b_i = 1$) and \emph{non-member} canaries ($b_i = 0$). Specifically, membership scores are used to compute the area under the receiver operating characteristic curve (ROC AUC), which quantifies the attack's ability to distinguish between reference canaries included in and excluded from $D$. We denote the resulting ROC AUC of MIAs on $\textit{LM}$ as $\tilde{\phi}$. 

\subsection{Experimental setup}
\label{sup:exp_setup}

We then measure how a target LLM memorizes across fuzzy duplicates by comparing MIA performances reached for fuzzy duplicates relative to the performance achieved for exact duplicates. The degree to which exact duplicates are memorized by a target LLM depends on the exact experimental setup $\mathcal{S}$, which we define by the following characteristics:

\begin{enumerate}
    \item The set of reference canaries $X_{\text{ref}}^i$ used as members and non-members. Throughout this work, we consider synthetically generated sequences generated using the pretrained Llama-2 7B~\cite{touvron2023llama2} as reference model $\textit{LM}_{\text{ref}}$. Unless stated otherwise, we
    use temperature $\mathcal{T}=1.0$. 
    \item The choice of pretrained model $\textit{LM}_0$, to be further pretrained on training dataset $D$ containing the fuzzy duplicates. 
    \item The hyperparameters used to further pretrain $\textit{LM}_0$ to get target model $\textit{LM}$. We keep this constant throughout this work, apart from the initial learning rate $\eta_0$. We opt to choose $\eta_0$ separately for each choice of pretrained model $\textit{LM}_0$, as we find this to be a crucial hyperparameter when comparing training progress and the absolute level of memorization across model size and architecture. 
    \item The MIA methodology and performance metric used to measure memorization. Throughout this work we use the ROC AUC computed on a balanced set of member and non-member reference canaries, and consider different MIA methods from the literature defined by their scoring function $\alpha$.
\end{enumerate}

We consider a range of models as the pretrained model $\textit{LM}_0$. In most experiments, we consider GPT-NEO 1.3B~\cite{gpt-neo} developed by EleutherAI. In further ablations, we also consider its smaller and larger versions, GPT-NEO 125M and 2.7B, and further consider three other models from different model families: Gemma-2B~\cite{team2024gemma} developed by Google, Phi-2~\cite{javaheripi2023phi} developed by Microsoft and Llama-3.2-1B~\cite{dubey2024llama} developed by Meta.  

In all of our experiments, we further pretrain $\textit{LM}_0$ on dataset with documents $D$ containing reference canaries and their fuzzy duplicates, which results in the target model \textit{LM}. We inject canaries in a collection of books available in the public domain, i.e. the original dataset $D_{\text{orig}}$. We use the open-source library~\cite{kpullygutenberg} to collect $\frac{C}{2}=100$ books made available under a permissive license on Project Gutenberg~\cite{projectgutenberg}. We collect books added to project Gutenberg after the release of GPT-NEO, which were thus likely not present in the original model's training dataset. The books we selected contain $8.8$M tokens (GPT-NEO) in total. For each reference canary $X_{\text{ref}}^i$ for which $b_i=1$, we inject the $n_{\text{dup}}=10$ fuzzy duplicates at random into one book and use this collection of modified books as dataset $D$ for continued pretraining. 

In all of our experiments we further pretrain $\textit{LM}_0$ on the modified books $D$ for $1$ epoch. We use as maximum sequence length $2048$ tokens, an effective batch size of $2$ and an Adam optimizer. We use a linear learning rater scheduler with as initial learning rate $\eta_0$ and weight decay $0.01$. Across pretrained models $\textit{LM}_0$, we empirically determine a value of $\eta_0$ to align with a training regime leading to a significant level of absolute memorization. We also show how the results vary with $\eta_0$ for one model GPT-NEO 1.3B (Section~\ref{sup:ablations}). 

For reference, further pretraining GPT-NEO 1.3B on $D$ takes roughly $2$ GPU-hours on A100 NVIDIA GPUs, with minor variations for other models. 

\subsection{Computing the exact duplicate equivalent} 
\label{sup:computing_rho}

We now detail the computation of the exact duplicate equivalent, $\rho$, for a given set of fuzzy duplicates generated by algorithm $\mathcal{A}$. First, we determine $\nu_{\text{eq}}$, which represents the equivalent number of exact repetitions that yield the same MIA AUC as observed for the fuzzy duplicates ($\tilde{\phi}$), under the same experimental setup $\mathcal{S}$. We then normalize $\nu_{\text{eq}}$ by the total number of fuzzy duplicates considered, ensuring that $\rho$ reflects the equivalent number of exact duplicates represented by a single fuzzy duplicate generated by $\mathcal{A}$.

To compute $\nu_{\text{eq}}$, we construct a modified training dataset $D_{\nu}$, where $\nu \in \{1, \ldots, n_{\text{dup}}\}$ exact copies of each reference canary $X_{\text{ref}}^i$ are injected. We use the same set of random variables $\{b_i\}$ generated during the fuzzy duplicate experiments. Formally, the modified dataset $D_{\nu}$ is defined as:

\begin{equation}
D_{\nu} = D_{\text{orig}} \cup \left( \bigcup_{i=1}^{C} \left( \{X_{\text{ref}}^i\}^{\nu} \cdot b_i \right) \right),
\end{equation}

where $\{X_{\text{ref}}^i\}^{\nu}$ denotes $\nu$ exact copies of $X_{\text{ref}}^i$. For each $\nu$, we train $\textit{LM}_0$ on $D_{\nu}$ to get target model $\textit{LM}_{\nu}$. We measure the performance of the MIA applied $\textit{LM}_{\nu}$ and denote this as $\phi_{\nu}$. 

\begin{figure}[t]
\centering
\includegraphics[width=0.45\linewidth]{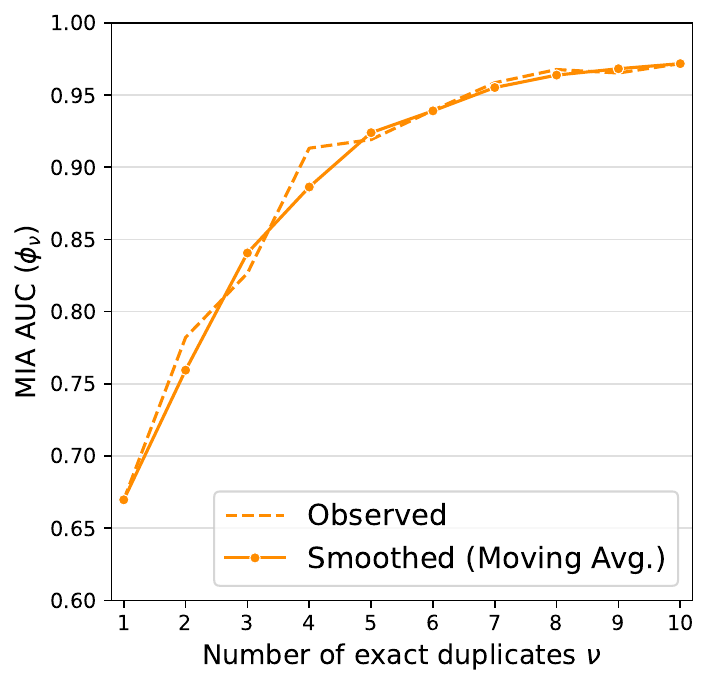}  
    \caption{\textbf{Smoothening the MIA AUC $\phi_{\nu}$ versus number of exact repetitions $\nu$.} Results for GPT-NEO 1.3B~\cite{gpt-neo} as $\textit{LM}_0$, initial learning rate $\eta=2e-6$, \textit{Ratio}~\cite{carlini2021extracting} as MIA methodology and reference canaries generated using temperature $\mathcal{T}=1.0$.} 
\label{fig:metric_smoothing}
\end{figure}

Figure~\ref{fig:metric_smoothing} illustrates the observed values of $\phi_{\nu}$ for increasing value of $\nu$ for GPT-NEO 1.3B~\cite{gpt-neo} as $\textit{LM}_0$, initial learning rate $\eta_0=2e-6$, \textit{Ratio}~\cite{carlini2021extracting} as MIA methodology and reference canaries generated using temperature $\mathcal{T}=1.0$. 

We then leverage the values for $\phi_{\nu}$ to compute $\nu_{\text{eq}}$ for a given $\tilde{\phi}$. To avoid that noise in the observed values impacts the comparison of computed values of $\nu_{\text{eq}}$, we smoothen the curve for exact duplicates using a moving average with window size of $3$. Figure~\ref{fig:metric_smoothing} shows how the smoothed curve compares to the observed values, confirming the changes to be minimal. Throughout this work, we leverage the smoothened curve for $\phi_{\nu}$.

For a given memorization level of fuzzy duplicates $\tilde{\phi}$, $\nu_{\text{eq}}$ is determined as the value of $\nu$ for which $\tilde{\phi} \approx \phi_{\nu_{\text{eq}}}$. We compute $\nu_{\text{eq}}$ through piece-wise, linear interpolation, or 

\begin{equation}
    \nu_{\text{eq}} = \nu' + \frac{\tilde{\phi} - \phi_{\nu'}}{\phi_{\nu' + 1} - \phi_{\nu'}},
\label{eq:n_eq}
\end{equation}

where $\nu'$ represents the number of exact duplicates such that $\phi_{\nu'} \leq \tilde{\phi} \leq \phi_{\nu' + 1}$.

\begin{figure}[!t]
\centering
\includegraphics[width=0.75\linewidth]{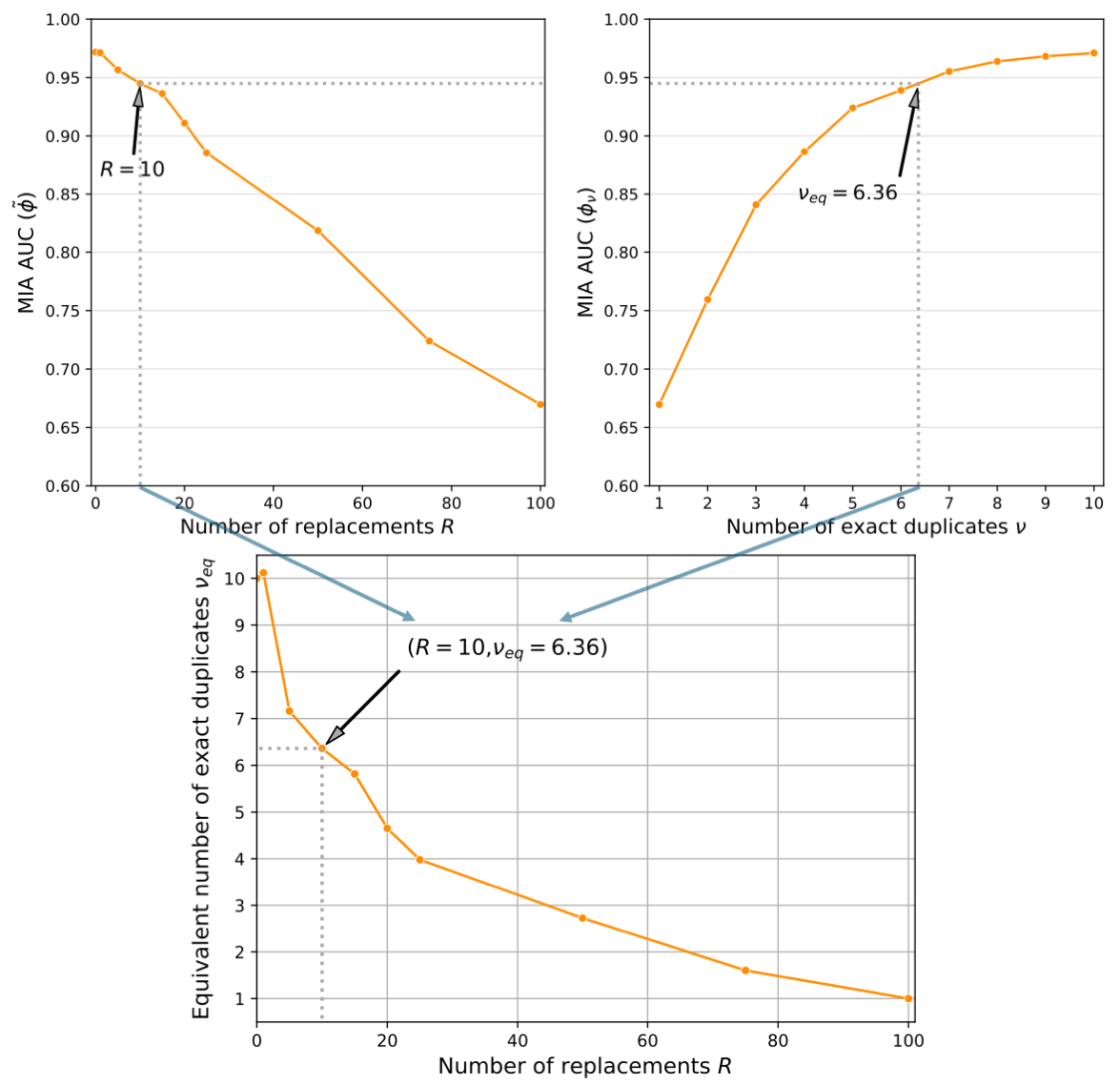}  
    \caption{\textbf{Computing the equivalent number of exact duplicates ($\nu_{\text{eq}}$).} We first compute the MIA performance (AUC) for fuzzy duplicates with number of replacements $R$, as well as the AUC for a range of exact duplicate counts. We then interpolate to find the equivalent exact duplicate count achieving similar MIA performance in the same experiment setup.} 
\label{fig:metric}
\end{figure}

Figure~\ref{fig:metric} illustrates how for $R=10$, we reach $\nu_{\text{eq}}=6.36$. We then repeat the same process for each observation $(R, \tilde{\phi})$, to quantify how $\nu_{\text{eq}}$ evolves with varying number of replacements $R$ across fuzzy duplicates. Note that, in some cases, we empirically find $\tilde{\phi}$ to reach values slightly larger than $\phi_{\nu=n_{\text{dup}}}$. We then also compute $\phi_{\nu}$ for values $\nu>n_{\text{dup}}$ until we find a $\nu_\text{max}$ for which $\tilde{\phi} \leq \phi_{\nu_\text{max}}$ and only then apply the interpolation from Equation~\ref{eq:n_eq}.

Finally, we want a value reflecting the degree to which a fuzzy duplicate constructed by $\mathcal{A}$ contributes to the memorization of the reference canary, independent of the number of fuzzy duplicates ($n_{\text{dup}}$) we inject in the training dataset $D$. Therefore, we normalize $\nu_{\text{eq}}$ to get the exact duplicate equivalent, $\rho$:

\begin{equation}
    \rho = \frac{\nu_{\text{eq}} - 1}{{n_\text{dup}} - 1}.
\label{eq:rho}
\end{equation}

Note that we need to subtract by $1$ in both the numerator and denominator, as we always include $1$ exact duplicate of the reference canary, or $X_1^i=X_{\text{ref}}^i$.

\subsection{Ablation studies} 
\label{sup:ablations}

It is well known that the extent to which LLMs memorize depends on many factors of the experimental setup $\mathcal{S}$, including for instance training hyperparameters and model size~\cite{carlini2022quantifying}. We refer to this as the \emph{absolute} level of memorization of the target model. Exactly for this reason, we have studied the mosaic memory throughout this work by computing the exact duplicate equivalent $\rho$, which captures the level of memorization \emph{relative} to the absolute level reached for the specific experimental setup. To further study whether our quantification of the mosaic memory of LLMs holds for different levels of absolute memorization, we here vary the core elements of the experimental setup $\mathcal{S}$ and evaluate how $\rho$ varies with the number of replacements $R$ in $\mathcal{A}_\text{replace}$ across $\mathcal{S}$.

As a baseline, we use the results for GPT-NEO 1.3B~\cite{gpt-neo} from Figure~\ref{fig:neq_vs_R_all_models}. Specifically, we consider reference canaries generated with temperature $\mathcal{T}=1.0$, and fuzzy duplicates generated using $\mathcal{A}_\text{replace}$ making \texttt{not evenly spread + not consistent} token replacements using the masked language model $\textit{MLM}$ and top $k=10$.

\textbf{Results across MIA methodology.} First, we compute the value of $\rho$ versus number of replacements $R$ when we use a different MIA methodology. We consider the three MIA methods described above: \textit{Loss}~\cite{yeom2018privacy}, \textit{Ratio}~\cite{carlini2019secret} using the pretrained Llama-2 7B~\cite{touvron2023llama2} as reference model and \textit{Min-K\% Prob}~\cite{shi2023detecting}.

Table~\ref{tab:main_results} reports the observed MIA AUC and the corresponding exact duplicate equivalent $\rho$ across replacements $R$ across MIA methodologies. We find $\rho$ to decrease similarly with increasing number of replacements $R$, independent of how the MIA AUC is computed or its absolute values. For instance, when $R=20$ replacements are being made across fuzzy duplicates, we find an exact duplicate equivalent $\rho$ of $0.401$, $0.406$ and $0.391$ for the attacks \textit{Loss}, \textit{Ratio} and \textit{Min-K\% Prob}, respectively. 

This allows us to use $\rho$ as a metric to study the mosaic memory of LLMs (e.g. for various algorithms $\mathcal{A}$ to generate fuzzy duplicates), independent of the absolute level of memorization. As we find the \emph{Ratio} MIA~\cite{carlini2021extracting} to consistently lead to higher AUC values across setups, we consider this methodology further in this work. 

\begin{table*}[h!]
    \centering
    \begin{tabular}{ccc|cc|cc}
    \toprule
         & \multicolumn{6}{c}{MIA methodology} \\
         & \multicolumn{2}{c}{\textit{Loss}} & \multicolumn{2}{c}{\textit{Ratio}} & \multicolumn{2}{c}{\textit{Min-K\% Prob}}  \\
         R & AUC & $\rho$ & AUC & $\rho$ & AUC & $\rho$ \\
        \hline
        \midrule
        $0$ & $0.93$ & $1$ & $0.97$ & $1$ & $0.94$ & $1$ \\ 
        \midrule
        \midrule
        $1$ & $0.93$ & $0.988$ & $0.97$ & $1.013$ & $0.95$ & $1.038$ \\ 
        \midrule
        $5$ & $0.92$ & $0.767$ & $0.96$ & $0.684$ & $0.95$ & $0.996$ \\ 
        \midrule
        $10$ & $0.89$ & $0.629$ & $0.94$ & $0.596$ & $0.92$ & $0.584$ \\ 
        \midrule
        $15$ & $0.87$ & $0.540$ & $0.94$ & $0.536$ & $0.91$ & $0.534$ \\ 
        \midrule
        $20$ & $0.84$ & $0.401$ & $0.91$ & $0.406$ & $0.88$ & $0.391$ \\ 
        \midrule
        $25$ & $0.83$ & $0.374$ & $0.89$ & $0.331$ & $0.87$ & $0.339$ \\ 
        \midrule
        $50$ & $0.75$ & $0.204$ & $0.82$ & $0.192$ & $0.81$ & $0.212$ \\ 
        \midrule
        $75$ & $0.69$ & $0.098$ & $0.72$ & $0.067$ & $0.74$ & $0.100$ \\
        \midrule
        \midrule
        $100$ & $0.64$ & $0$ & $0.67$ & $0$ & $0.68$ & $0$ \\ 
        \midrule
        \bottomrule
    \end{tabular}
    \caption{\textbf{Quantifying the mosaic memory of GPT-NEO 1.3B across MIA methodologies.} MIA AUC and the exact duplicate equivalent $\rho$ for fuzzy duplicates (varying number of replacements R using $\mathcal{A}_\text{replace}$) across MIA methodologies. Results for GPT-NEO 1.3B~\cite{gpt-neo} as $\textit{LM}_0$, initial learning rate $\eta_0=2e-6$ and reference canaries generated using temperature $\mathcal{T}=1.0$.}
    \label{tab:main_results}
\end{table*}

\begin{figure*}
    \centering
    \subfloat[]
    {\includegraphics[width=0.6\textwidth]{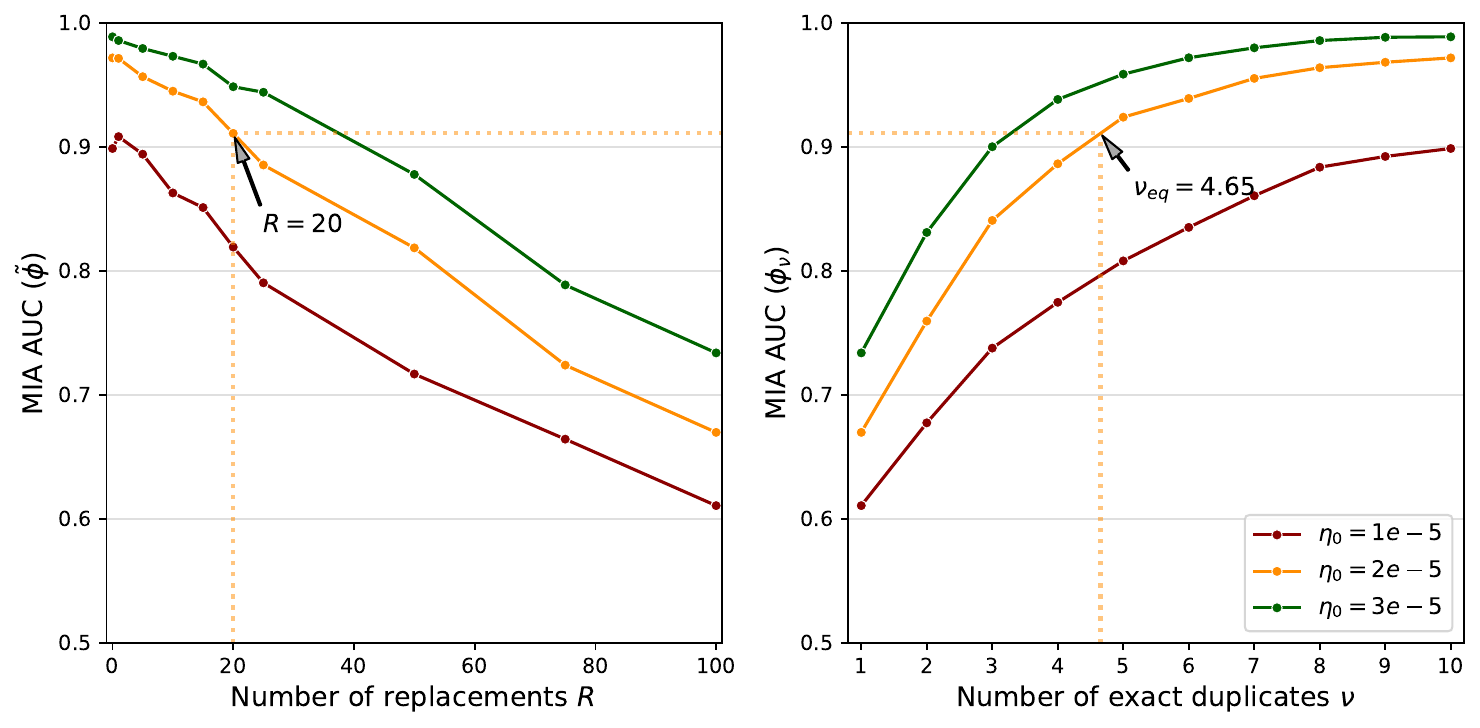}}
    \qquad
    \subfloat[]
    {\includegraphics[width=0.3\textwidth]{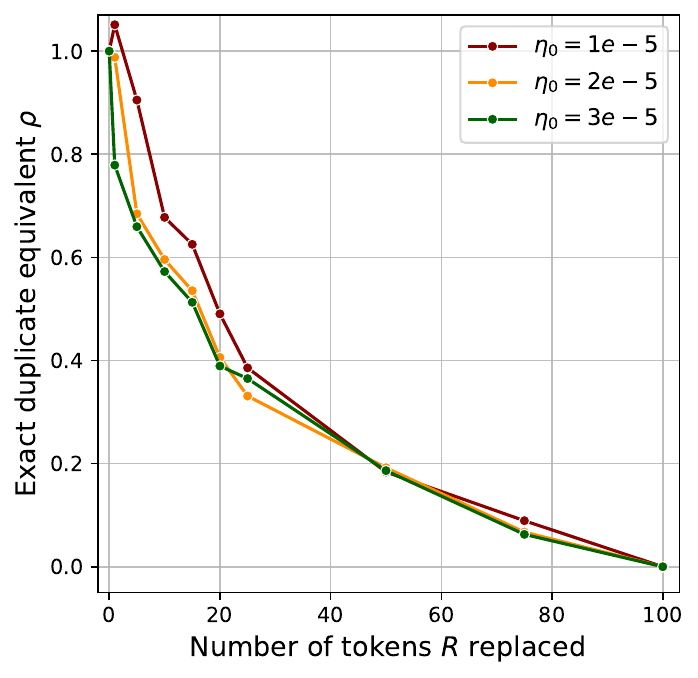}}
    \caption{\textbf{Varying the learning rate.} MIA AUC and equivalent number of exact duplicates ($\nu_{\text{eq}}$) for varying initial learning rate $\eta_0$ (a) with the corresponding exact duplicate equivalent $\rho$ (b). Results for GPT-NEO 1.3B~\cite{gpt-neo} as $\textit{LM}_0$  and reference canaries generated using temperature $\mathcal{T}=1.0$.}
    \label{fig:ablation_learning_rate}
\end{figure*}

\textbf{Results across learning rate.} Figure~\ref{fig:ablation_learning_rate} illustrates how GPT-NEO 1.3B memorizes fuzzy duplicates across different initial learning rates ($\eta_0$). As shown in Figure~\ref{fig:ablation_learning_rate}(a), the MIA AUC (measuring the absolute memorization) increases with $\eta_0$, achieving near-perfect performance (AUC of 1.0) for sequences with more than $6$ exact duplicates when $\eta_0=3e-5$. This aligns with expectations: larger learning rates result in more substantial updates to model weights when a sequence is encountered during training, making the sequence's presence in the training data more detectable by the MIA.

We then continue to compute the exact duplicate equivalent $\rho$. Figure~\ref{fig:ablation_learning_rate}(b) demonstrates that the progression of $\rho$ with respect to the number of replaced tokens ($R$) remains consistent across different values of $\eta_0$. This indicates that our relative memorization metric, $\rho$, is robust to variations in absolute memorization levels, enabling comparisons of LLM memorization across settings. Furthermore, this consistency suggests that LLMs exhibit a mosaic memory, i.e. effectively memorizing fuzzy duplicates, in a similar manner across varying hyperparameter choices, such as learning rate.

\begin{figure*}
    \centering
    \subfloat[]
    {\includegraphics[width=0.6\textwidth]{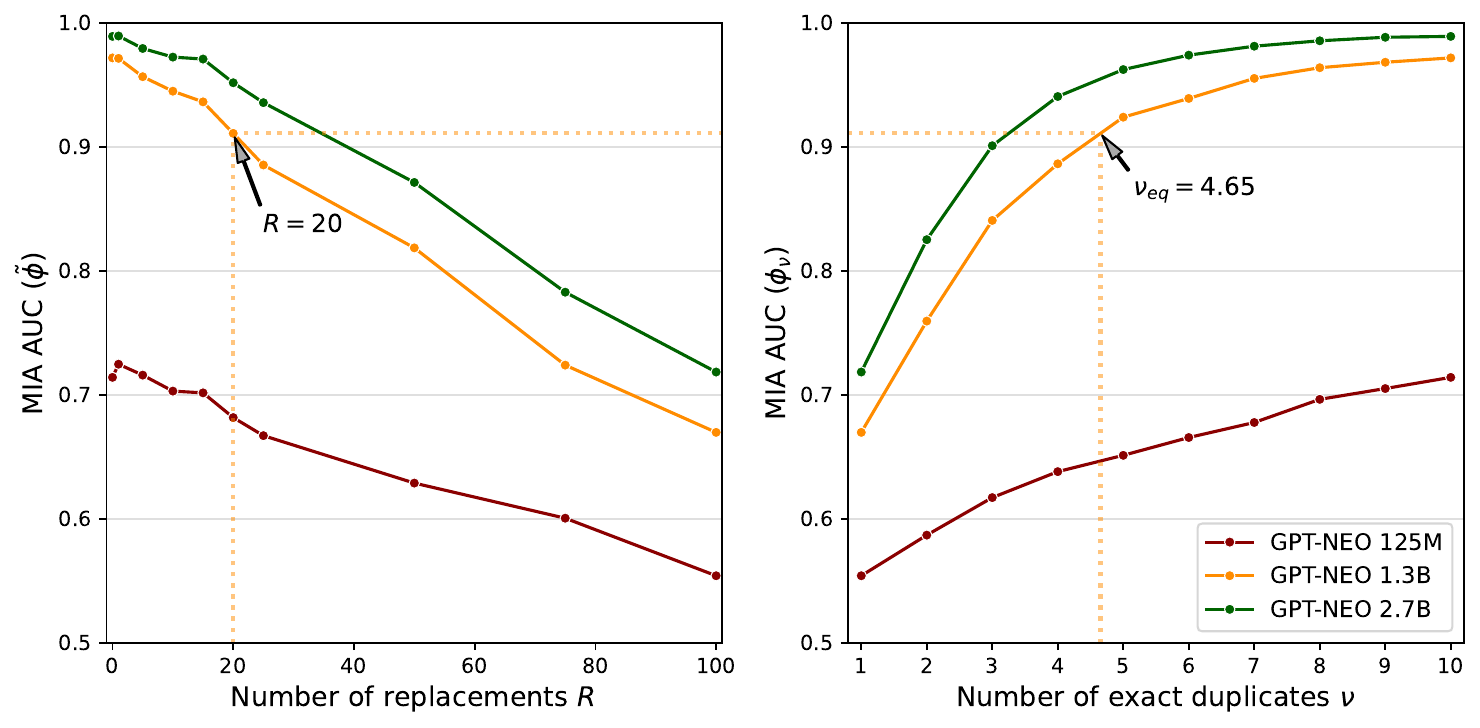}}
    \qquad
    \subfloat[]
    {\includegraphics[width=0.3\textwidth]{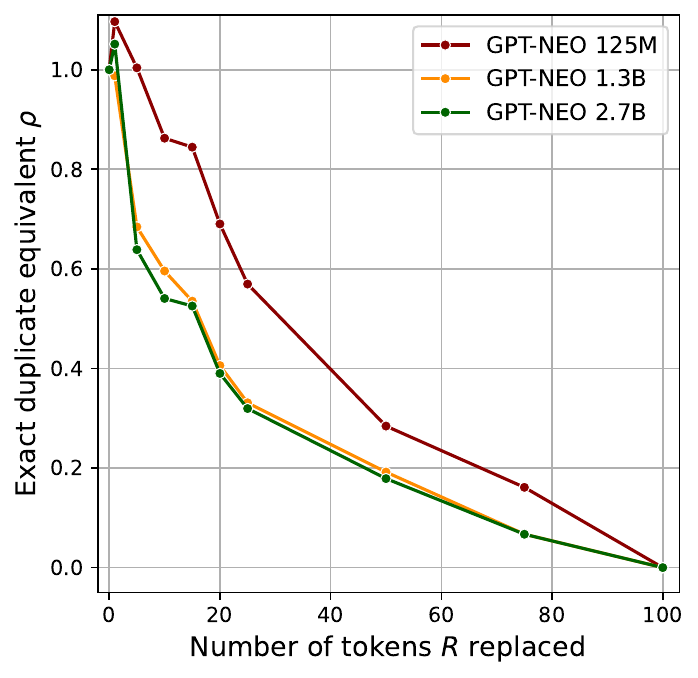}}
    \caption{\textbf{Varying the model size.} MIA AUC and equivalent number of exact duplicates ($\nu_{\text{eq}}$) for varying model size of target model GPT-NEO~\cite{gpt-neo} (a) with the corresponding exact duplicate equivalent $\rho$ (b). Results for initial learning rate $\eta_0=2e-6$ and reference canaries generated using temperature $\mathcal{T}=1.0$.}
    \label{fig:ablation_model_size}
\end{figure*}

\textbf{Results across model size.} We repeat the same experiment while keeping the initial learning rate $\eta_0=2e-5$, yet now varying the size of the target LLM. Specifically, we also consider the smaller and larger GPT-NEO models~\cite{gpt-neo}, of 125 million and 2.7 billion parameters, respectively. Figure~\ref{fig:ablation_model_size}(a), shows that the absolute level of memorization strongly depends on the model size. For instance, GPT-NEO 125M only reaches the MIA AUC of $0.71$ at $\nu=10$ exact duplicates, significantly lower than the near-perfect AUC reached for GPT-NEO 2.7B in the exact same setup. This again aligns with expectations: a model with more parameters has more capacity to memorize specific sequences, as also consistent with the literature~\cite{carlini2022quantifying}. 

We further examine the relative memorization of fuzzy duplicates ($\rho$). Our results indicate a similar mosaic memory across GPT-NEO 1.3B and GPT-NEO 2.7B models. Interestingly, GPT-NEO 125M exhibits slightly higher relative memorization, suggesting smaller models may rely more strongly on syntactic patterns, memorizing fuzzy duplicates proportionally more compared to their larger counterparts.

\begin{figure*}
    \centering
    \subfloat[]
    {\includegraphics[width=0.6\textwidth]{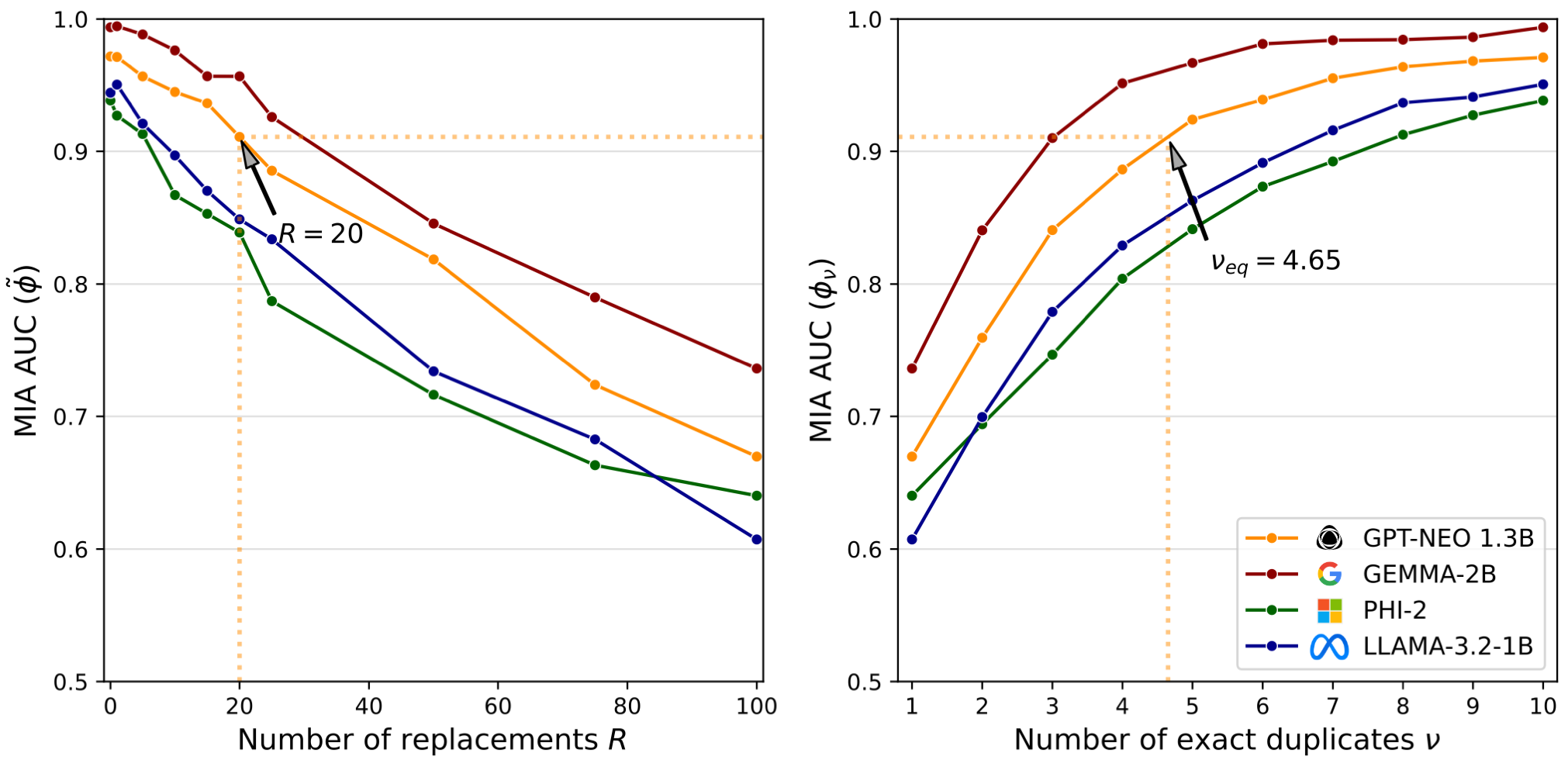}}
    \qquad
    \subfloat[]
    {\includegraphics[width=0.3\textwidth]{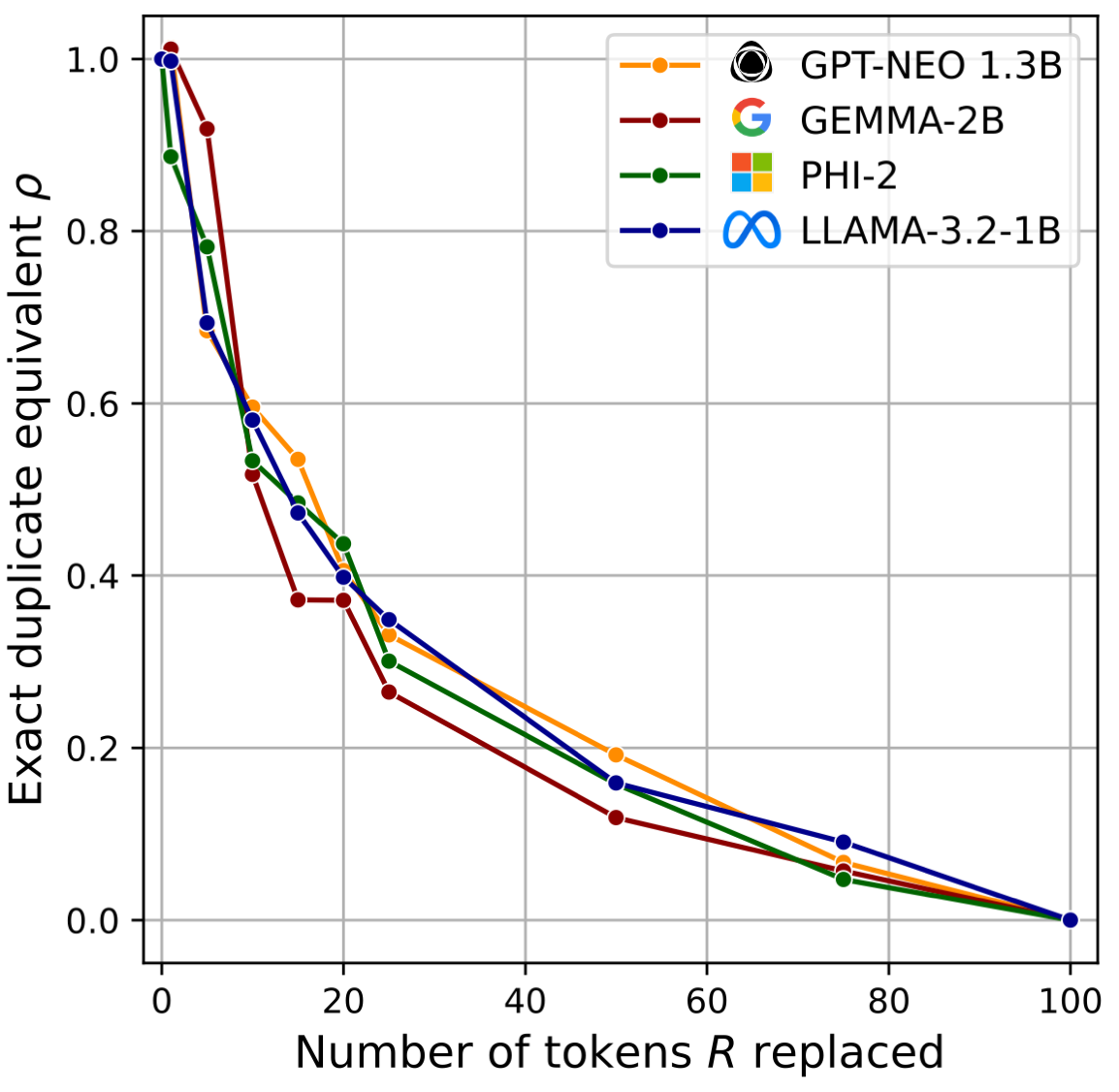}}
    \caption{\textbf{Varying the model family.} MIA AUC and equivalent number of exact duplicates ($\nu_{\text{eq}}$) for varying choice of pretrained model $\textit{LM}_0$ (a) with the corresponding exact duplicate equivalent $\rho$ (b). Results for initial learning rates from Table~\ref{tab:learning_rates} and reference canaries generated using temperature $\mathcal{T}=1.0$.}
    \label{fig:ablation_model_family}
\end{figure*}

\textbf{Results across model families.} Figures~\ref{fig:neq_vs_R_all_models} and~\ref{fig:ablation_model_family} illustrates how $\rho$ evolves with number of replacements $R$ across all pretrained models $\textit{LM}_0$ considered in this work~\cite{gpt-neo,team2024gemma,javaheripi2023phi,dubey2024llama}. Across model families, we empirically found a distinct impact of choice of initial learning rate $\eta_0$ and the absolute level of MIA performance and memorization. Table~\ref{tab:learning_rates} summarizes the values we empirically set to have a meaningful level of absolute memorization (MIA AUC $>0.5$). Our results above show how the value of $\rho$ is fairly consistent across different values of $\eta_0$.

\begin{table*}[h!]
    \centering
    \begin{tabular}{cc}
    \toprule
        Model $\textit{LM}_0$ & Initial learning rate $\eta_0$ \\
        \hline
        \midrule
        GPT-NEO \{125M, 1.3B, 2.7B\}~\cite{gpt-neo} & $2e-6$\\
        Gemma-2B~\cite{team2024gemma}  & $1e-5$ \\
        Phi-2~\cite{javaheripi2023phi} & $5e-5$ \\
        Llama-3.2-1B~\cite{dubey2024llama} & $2e-6$ \\
        \midrule
        \bottomrule
    \end{tabular}
    \caption{Initial learning rates $\eta_0$ used for continued pretraining across pretrained models $\textit{LM}_0$.}
    \label{tab:learning_rates}
\end{table*}

\textbf{Results across temperature for the reference canaries.} Thus far, we have only considered one set of reference canaries, namely the ones generated by sampling from the reference model $\textit{LM}_{\text{ref}}$ (Llama-2 7B~\cite{touvron2023llama2}) with temperature $\mathcal{T}=1.0$. We note that $\mathcal{T}=1.0$ is commonly used when generating synthetic text from pretrained LLMs, and we also empirically confirm that synthetic sequences generated using $\mathcal{T}=1.0$ lead to meaningful sequences (see Table~\ref{tab:reference_canaries_temp}). 

We now repeat the baseline experiment for GPT-NEO 1.3B~\cite{gpt-neo} as $\textit{LM}_0$ and initial learning rate $\eta_0=2e-6$ while varying temperature $\mathcal{T}$. As such, we can verify whether our results are consistent regardless of the reference canaries choice. 

\begin{figure}[!ht]
\centering
\includegraphics[width=0.45\linewidth]{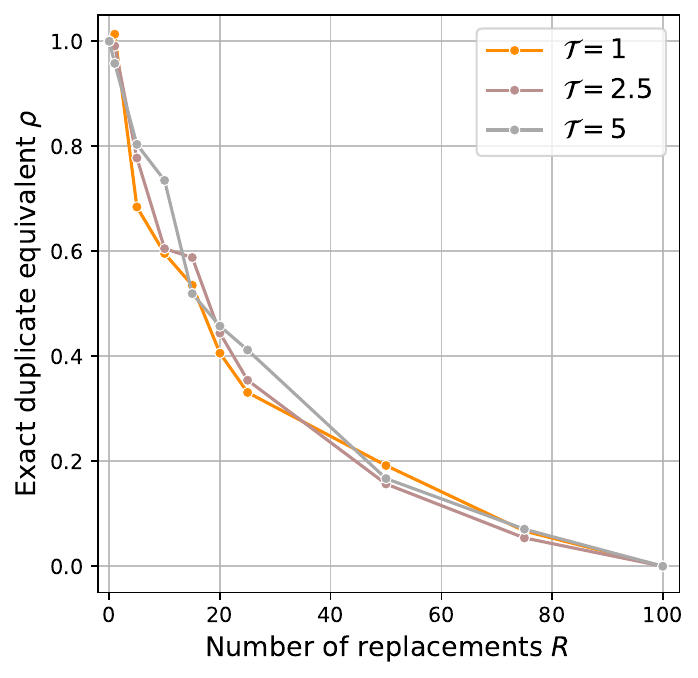}  
    \caption{\textbf{Varying the temperature $\mathcal{T}$.} The exact duplicate equivalent $\rho$ for fuzzy duplicates across number of replacements made, considering different values of the temperature $\mathcal{T}$ used to synthetically generate reference canaries $X_{\text{ref}}^i$ using Llama-2 7B~\cite{touvron2023llama2} as reference model $\textit{LM}_{\text{ref}}$.} 
\label{fig:ablation_temperature}
\end{figure}

Figure~\ref{fig:ablation_temperature} shows how $\rho$ decreases for increasing number of $R$ token replacements made across fuzzy duplicates, for different values of the temperature used to generate the reference canaries. We find that the memorization of fuzzy duplicates is strikingly consistent across a wide range of temperature values, even at $\mathcal{T}=5$.

This is especially remarkable, as we empirically find sequences generated with $\mathcal{T}=5$ to correspond to quite nonsensical sentences. These results thus suggest that LLMs memorize across fuzzy duplicates regardless of whether the reference canary is a coherent sequence of tokens. This, again, supports the thesis that an LLM's mosaic memory is predominantly syntactic rather than semantic.

\subsection{Ablations for fuzzy duplicate generation techniques} 
\label{sup:ablations_position} 

In this section we consider distinct ways of creating fuzzy duplicates not discussed in the main body of the paper.  

\textbf{Information loss.} Token replacement strategy $\mathcal{A}_\text{replace}$ discussed before inevitably leads to an \emph{information loss}, as when tokens are replaced, the modes sees fewer instances of the correct tokens. We here investigate the impact of information loss on memorization, separately from other properties of $\mathcal{A}_\text{replace}$. For that we consider various fuzzy duplicate generation strategies, all of which maintain the same level of information loss across fuzzy duplicates.

We distinguish the following ways in which a fuzzy duplicate can differ by $R$ tokens from the tokenized reference canary $T(X_{\text{ref}}^i)$: 

\begin{enumerate}
    \item Removing $R$ tokens at the end of the sequence (\emph{suffix}). We here consider all fuzzy duplicates $\{X_j^i \mid j=2, \ldots, n_{\text{dup}}\}$ to be equal to the first $|T(X_{\text{ref}}^i)| - R$ tokens of the reference canary. 
    \item Removing $R$ tokens at the start of the sequence (\emph{prefix}). We here consider all fuzzy duplicates $\{X_j^i \mid j=2, \ldots, n_{\text{dup}}\}$ to be equal to the last $|T(X_{\text{ref}}^i)| - R$ tokens of the reference canary. 
    \item Removing $R$ tokens randomly. We here remove the same $R$ random tokens for all fuzzy duplicates $\{X_j^i \mid i=2, \ldots, n_{\text{dup}}\}$. The tokens to be removed are selected to be evenly distributed across the reference canary (\texttt{evenly spread + consistent}, see Section~\ref{sup:fuzz_dup}). This ensures that the subsequences between the replaced tokens are of equal length. 
    \item Replacing $R$ tokens. We here replace $R$ tokens by sampling another token from top $k=10$ tokens predicted by the $\textit{MLM}$. Again, the tokens to be removed are selected to maximize the length of the overlapping subsequences and are consistent across fuzzy duplicates (\texttt{evenly spread + consistent}, see Section~\ref{sup:fuzz_dup}). The random tokens used for replacement differ across fuzzy duplicates. 
\end{enumerate}

\begin{figure}[t]
    \centering
    \subfloat[]
    {\includegraphics[width=0.45\textwidth]{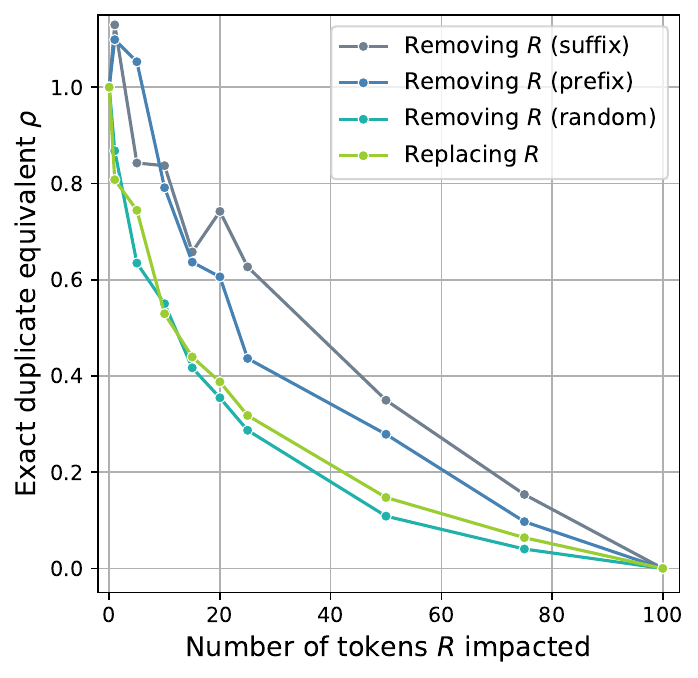}}
    \qquad
    \subfloat[]
    {\includegraphics[width=0.45\textwidth]{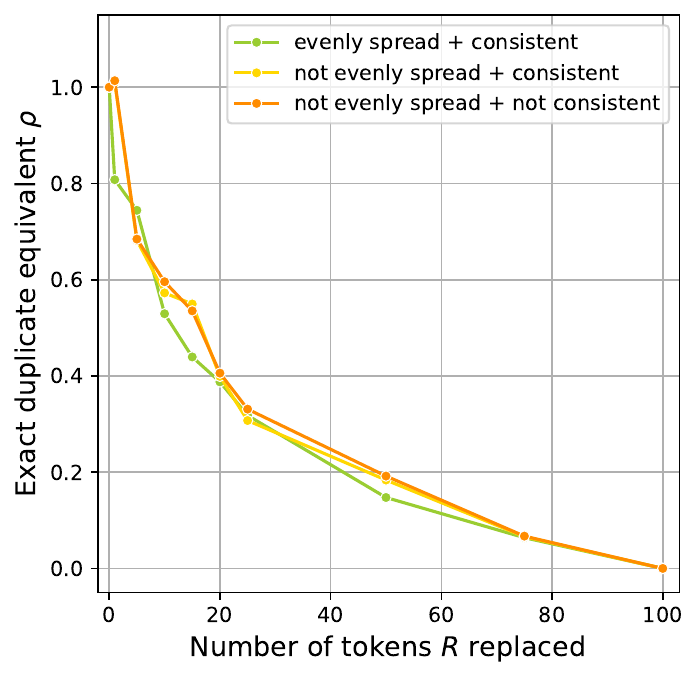}}
    \caption{\textbf{Ablations for fuzzy duplicate generation techniques.} (a) The exact duplicate equivalent $\rho$ for fuzzy duplicates when tokens are removed (randomly or from the prefix or suffix) or replaced with a semantically meaningful token. (b) The exact duplicate equivalent $\rho$ for fuzzy duplicates when tokens are replaced with a semantically meaningful token, varying which tokens in the reference canary are replaced.}
    \label{fig:ablations_position}
\end{figure}

Figure~\ref{fig:ablations_position}(a) shows how the exact duplicate equivalent $\rho$ changes with the number of impacted tokens $R$ for the four kinds of information loss. 

First, we find that removing the tokens from the prefix or suffix of the sequence maintains the most memorization for increasing $R$. This suggests that most information is retained by the model when the exact overlap between the reference canary and the fuzzy duplicates remains the largest, i.e. equal to $|T(X_{\text{ref}}^i)| - R$. Notably, removing tokens from the suffix leads to slightly higher values of $\nu_{\text{eq}}$. We hypothesize this is because the MIA score is computed for the full length reference canary (i.e. $\alpha(X_{\text{ref}}^i)$, see Section~\ref{sup:mias}). When tokens are removed only from the canary suffix instead of from its prefix, this means that the context used for token-level predictions is more consistent between model training and inference during the MIA, facilitating membership inference.  

When the removed tokens are spread across the sequence, the memorization ($\rho$) drops quite sharply, yet remains significantly higher than the baseline memorization of $\rho=0$. For instance, when $R=20$ random tokens are removed (and the maximum overlapping subsequence between fuzzy duplicates and the reference canary is no more than $4$ tokens), the exact duplicate equivalent still remains as high as $\rho=0.4$. These results show that the target LLM's memorization is able to connect across missing pieces. Lastly, we compare this to when tokens are not removed but replaced by a semantically meaningful token, finding a slight increase in $\rho$ for larger values of $R$. We hypothesize that removing tokens introduces additional noise compared to maintaining the coherence of the sequence by replacement, making it slightly harder for the LLM to memorize across fuzzy duplicates. 

\textbf{Token selection.} Recall from Section~\ref{sup:fuzz_dup} that when replacing $R$ tokens, we distinguished between replacements evenly and not-evenly distributed within the reference canary and consistently and not-consistently across fuzzy duplicates. We now examine the corresponding impact on memorization.

Recall that above (Figure~\ref{fig:ablations_position}(a)), we replace $R$ tokens evenly distributed within the reference canary and choose to replace the exact same tokens across fuzzy duplicates (\texttt{evenly spread + consistent}). We now also consider the cases in which the tokens to be replaced are not evenly distributed, both with the replaced tokens consistent across fuzzy duplicates (\texttt{not evenly spread + consistent}) and not consistent (\texttt{not evenly spread + not consistent}).

Figure~\ref{fig:ablations_position}(b) illustrates that the choice of token replacement positions has limited impact on the memorization of the fuzzy duplicates. Indeed, $\rho$ drops very similarly with increasing value of $R$ across all three scenarios. We do see a slight increase of $\rho$ when the tokens to be replaced are not uniformly spread across the reference canary. We hypothesize that the maximum overlapping subsequence between the reference canary and the fuzzy duplicate to be slightly larger in this case, which would help memorization. Notably, $\rho$ remains similarly high when the replaced tokens are not consistent across fuzzy duplicates, which shows how the target LLM memorizes across distinctly overlapping fragments of sequences.  

\subsection{Generating paraphrases as fuzzy duplicates} 
\label{sup:paraphrases}

In the results from Table~\ref{tab:paraphrases_results}, we use $\mathcal{A}_{paraphrase}$ to construct the fuzzy duplicates $\{X_j^i \mid j=2, \ldots, n_{\text{dup}}\}$; i.e. we query an instruction-tuned LLM for $n_{\text{dup}} - 1$ paraphrases of the reference canary $X_{\text{ref}}^i$. Specifically, we query the models Meta-Llama-3-8B-Instruct~\cite{llama3modelcard}, Mistral-7B-Instruct-v0.2~\cite{jiang2023mistral} and GPT-4o~\cite{openai2024gpt4o} using the following system and user prompts:

\begin{enumerate}
    \item \textbf{System prompt:} \textit{"You are an assistant tasked with rephrasing the provided text in 9 different ways. Keep the original meaning intact (including the original natural or code language), but rephrase each version as if you are replacing the sentence entirely. Number the rephrased sequences using 1. to 9. and separate each by '---', like this: '1. rephrase 1 --- 2. rephrase 2 --- .. --- 9. rephrase 9'."}
    \item \textbf{User prompt:} \textit{"Can you rephrase the following sequence? \{$X_{\text{ref}}^i$\}"}
\end{enumerate}

We generate the answer from the model with a maximum of new tokens of $1024$ and temperature of $0.6$ while sampling consecutively from the top predicted tokens for which the total predicted probability sums up to $0.9$ (except for GPT-4o~\cite{openai2024gpt4o} for which we did not control the sampling parameters). We query the model until the processed output leads to $n_{\text{dup}} - 1$ distinct paraphrases. We then repeat the same process as in the main experiment, finetuning the target model on a dataset always containing $X_{\text{ref}}^i$ and now its $n_{\text{dup}} - 1$ paraphrases. Table~\ref{tab:paraphrases_examples} shows one example of rephrased fuzzy duplicate for each instruction-tuned model used. 

\begin{table}[t!]
    \centering
    \begin{tabular}{p{2cm}p{10cm}}
    \toprule
        \multicolumn{1}{p{2cm}}{Paraphrase model} & \multicolumn{1}{c}{Selected paraphrased fuzzy duplicate} \\
        \midrule
        \midrule
        None ($X_{\text{ref}}^i$) & \textit{A few years ago I came across a video of the great jazz drummer Art Blakey playing live. The drummer is playing at a ferocious pace yet with an exacting control. I've been practicing my drums more and more, and I wanted to be able to play with that same combination of precision and intensity. I decided to focus on improving my timing by playing a few exercises. The following exercises use quarter notes. The left hand plays with the metronome while the right hand keeps a stead} \\
        \midrule
        \midrule
        Llama-3-8B~\cite{llama3modelcard} & \textit{It was Art Blakey's electrifying live performance that first drew me to the world of jazz drumming, and I've been striving to replicate his unique blend of precision and power ever since.} \\
        \midrule
        Mistral-7B~\cite{jiang2023mistral} & \textit{About a decade ago, I encountered a video of the iconic jazz drummer Art Blakey performing live. Blakey showcased a breathtaking combination of lightning-fast beats and unwavering control. Driven to replicate that performance style, I've been intensifying my drum practice sessions. To perfect my timing, I've selected a few exercises.} \\
        \midrule
        GPT-4o~\cite{openai2024gpt4o} & \textit{A few years back, I discovered a live performance video of the renowned jazz drummer Art Blakey. His ability to play at a blazing speed while maintaining exact control fascinated me. As a result, I've been increasingly committed to practicing the drums, striving to emulate that precise intensity. I decided to concentrate on enhancing my timing through certain exercises. These exercises focus on quarter notes, where the left hand aligns with the metronome, and the right hand keeps a steady beat.} \\
        \midrule
        \bottomrule
    \end{tabular}
    \caption{Examples of paraphrased fuzzy duplicates ($\mathcal{A}_{paraphrase}$) across instruction-tuned LLMs.}
    \label{tab:paraphrases_examples}
\end{table}

Lastly, to put the results from Table~\ref{tab:paraphrases_results} in perspective, we also compute the $n$-gram overlap between the fuzzy duplicates and the reference canary for fuzzy duplicates constructed by token replacements ($\mathcal{A}_\text{replace}$). Specifically, we also compute the mean and standard deviation overlap in $n$-grams (in GPT-NEO tokens) for the results from Figure~\ref{fig:semantic_results} for $k=|\mathcal{V}_{\text{MLM}}|$. Note that we consider all possible $n$-grams in a sequence and that an $n$-gram can appear in the fuzzy duplicate also when it is not in the same location. The results are summarized in Table~\ref{tab:ngram_overlap_replace}. 

\begin{table*}[t]
    \centering
    \begin{tabular}{ccccc}
    \toprule
        Fuzzy duplicates & & \multicolumn{3}{c}{$n$-gram overlap} \\
        using $\mathcal{A}_\text{replace}$ & $\rho$ & $n=1$ & $n=2$ & $n=4$ \\
        \hline
        \midrule
        $R=5$ & $0.68$ & $97.83\pm1.23$ & $91.40\pm2.03$ & $78.90\pm2.86$\\
        \midrule
        $R=10$ & $0.58$ & $95.48\pm1.89$ & $83.37\pm3.49$ & $59.81\pm5.10$\\
        \midrule
        $R=15$ & $0.39$ & $93.18\pm2.53$ & $75.46\pm5.13$ & $40.47\pm5.79$\\
        \midrule
        $R=20$ & $0.35$ & $90.34\pm3.44$ & $66.59\pm6.78$ & $19.22\pm4.34$\\
        \midrule
        $R=25$ & $0.28$ & $87.63\pm4.17$ & $57.40\pm7.53$ & $1.07\pm0.29$\\
        \midrule
        \bottomrule
    \end{tabular}
    \caption{$n$-gram overlap for fuzzy duplicates constructed using  $\mathcal{A}_\text{replace}$ with $k=|\mathcal{V}_{\text{MLM}}|$ (results from Figure~\ref{fig:semantic_results}), to put the results from Table~\ref{tab:paraphrases_results} in perspective. }
    \label{tab:ngram_overlap_replace}
\end{table*}

\subsection{Collecting fuzzy duplicates from SlimPajama} 
\label{sup:slimapajam}

In this section we provide further technical details on how we have collected fuzzy duplicates present in SlimPajama, while also providing additional results. 

\textbf{Distance metric.} First, to define what constitutes as a fuzzy duplicate, we consider a range of potential metrics to compute the distance between sequences. A sequence is then considered as a fuzzy duplicate of a target sequence if both sequences are close according to this distance. 

Throughout most of the experiments presented in this work, we have introduced a predefined perturbation ($\mathcal{A}$) to the reference canary and measured its impact on memorization by computing the corresponding exact duplication equivalent $\rho$. Now, as we shift to analyzing a real-world dataset, we need a definition of fuzzy duplicates, i.e. a distance metric, that aligns with our prior experiments while also enabling us to estimate their impact on memorization in practical scenarios.

\begin{figure}[t]
\centering
\includegraphics[width=0.9\linewidth]{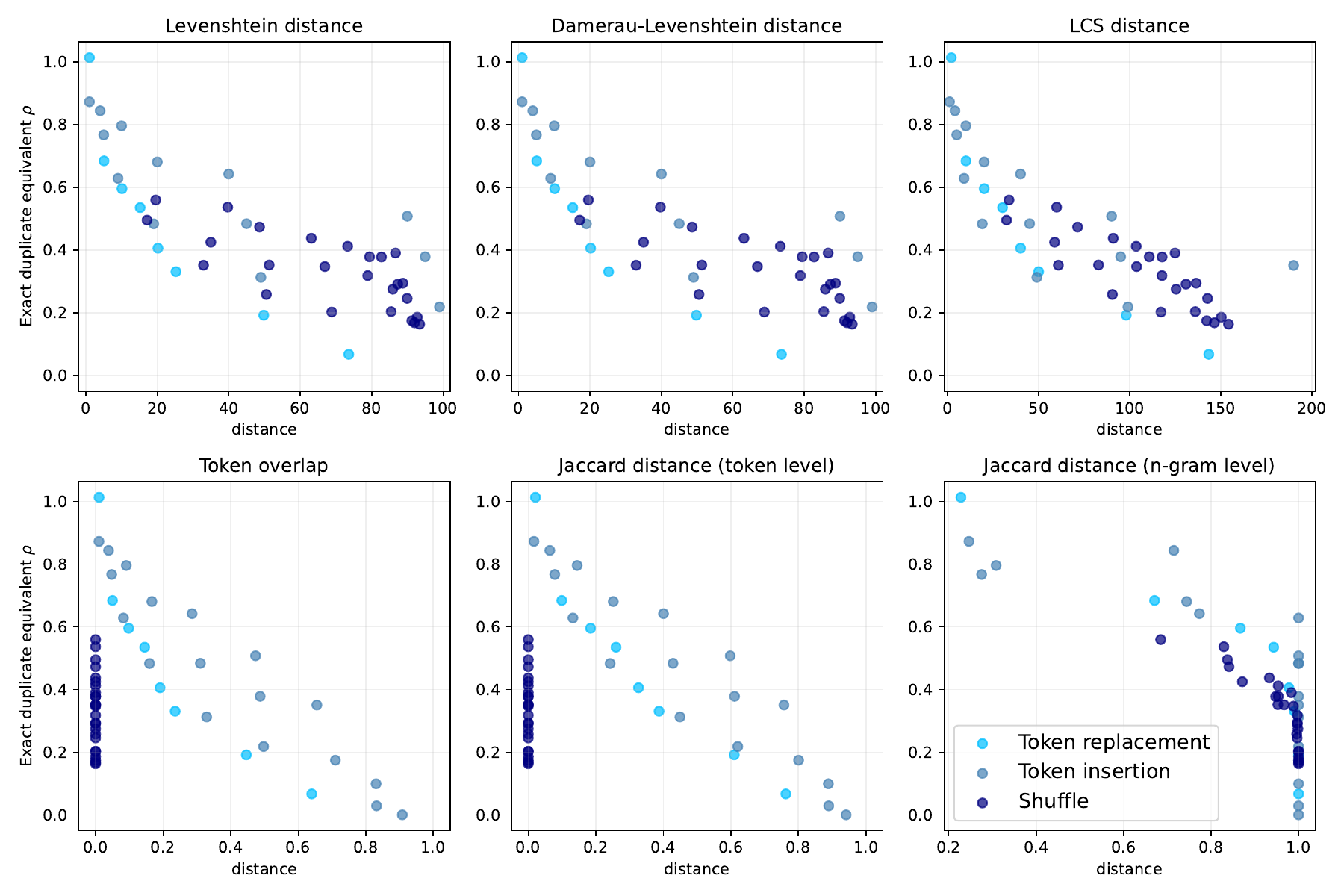} 
    \caption{Relationship between distance metrics and exact duplicate equivalent $\rho$ for token replacement, insertion, and shuffling experiments.} 
\label{fig:distances_scatterplot_all}
\end{figure}

To this end, we consider a range of metrics to compute the distance between two sequences of tokens, and show how well each of them relate to the value of $\rho$. We consider the results of three of our main experiments, where we estimated $\rho$ values for token replacement ($\mathcal{A}_\text{replace}$, Figure~\ref{fig:neq_vs_R_all_models}), insertion ($\mathcal{A}_\text{insert}$, Figure~\ref{fig:unraveling}(a)) and shuffling ($\mathcal{A}_\text{shuffle}$, Figure~\ref{fig:unraveling}(b)), and compute the target distance metric between the reference canary and its fuzzy duplicates. We then replace experiment-specific distance metrics used before (number of replaced tokens $R$, number of inserted tokens $X_\text{insert}$ and Kendall tau distance $\tau$ respectively) with the target distance metric and plot the results of all three experiments on one graph with the target distance as a shared x-axis (Figure~\ref{fig:distances_scatterplot_all}). 

We consider the following distance metrics:
\begin{enumerate}
    \item \textbf{Levenshtein distance}. This metric measures the minimum number of single-token operations (insertions, deletions, or substitutions) required to transform one sequence of tokens into another.
    \item \textbf{Levenshtein-Damerau distance}. This metric extends Levenshtein distance by allowing transposition of adjacent tokens as a single operation, making it better suited for capturing typographical errors.
    \item \textbf{Longest Common Subsequence (LCS) distance}. This metric calculates the length on the longest common subsequence (not necessarily consecutive) as the measure of similarity. The distance is then computed by substracting the LCS length from the input length.
    \item \textbf{Token overlap (multiset)}. This metric calculates one minus the intersection of tokens between two sequences divided by the maximum number of tokens in either sequence. As such, it represents how many tokens appear in both sequences while accounting for token frequency.
    \item \textbf{Jaccard distance (token level)}. This metric represents the dissimilarity between two sequences by calculating one minus the ratio of the size of token intersection to the size of token union.
    \item \textbf{Jaccard distance (n-gram level)}. This metric measures the dissimilarity between two sequences by calculating one minus the ratio of the size of n-gram (in tokens) intersection to the size of n-gram union, capturing sequence-level differences. 
\end{enumerate}

Figure~\ref{fig:distances_scatterplot_all} shows that Levenshtein, Levenshtein-Damerau and LCS distances show good correlation between the distance and the $\rho$ value across all three experiments, effectively capturing fuzzy duplicates that contribute to mosaic memory. Token overlap (multiset) and token-level Jaccard distance both perform well for token replacement and token insertions, but fail to capture shuffling -- as both metrics are order-independent, they assign zero distance to all shuffled fuzzy duplicates. N-gram-level Jaccard distance ($n$=13), on the other hand, suffers from the opposite problem: most sequence perturbations we have explored lead to very little overlap in 13-grams, with the Jaccard distance values collapsing to near 1 for a wide range of $\rho$ values. 

To identify fuzzy duplicates in a real-world dataset, we have chosen the Levenstein distance. This metric shows equally good correlation as other well-performing metrics, is conceptually simple and naturally encapsulates types of sequence perturbations we explored previously (token replacement, insertion and shuffling).

\textbf{Collecting duplicates.} We now estimate the prevalence of fuzzy duplicates as defined by our selected metric in a real-world dataset used for LLM training. As a dataset, we consider SlimPajama~\cite{cerebras2023slimpajama}, which contains 627 billion GPT-NEO tokens. We then consider a sequence a fuzzy duplicate from a target sequence if their Levenshtein distance is below a certain threshold. We hypothesize that a substantial portion of the dataset would have a large number of such fuzzy duplicates, thus heavily contributing to model memorization. 

First, we identify a set of target sequences for which we will seek fuzzy duplicates. These target sequences can be selected in various ways, and the exact selection criteria likely impacts the associated number of fuzzy duplicates. We hypothesize that the number of fuzzy duplicates identified for a certain target sequence has a positive correlation with the number of times this target sequence appears exactly in the dataset. Hence, we first group sequences by their number of exact repetitions in buckets to then search for fuzzy duplicates for target sequences sampled from each bucket. Note that we do not necessarily aim to show that \textit{all} sequences in the dataset have fuzzy duplicates. Rather, we seek to show that a significant portion of the dataset has a large number of fuzzy duplicates, thus raising memorization concerns.

\begin{figure*}[t]
    \centering
    \subfloat[]
    {\includegraphics[width=0.4\textwidth]{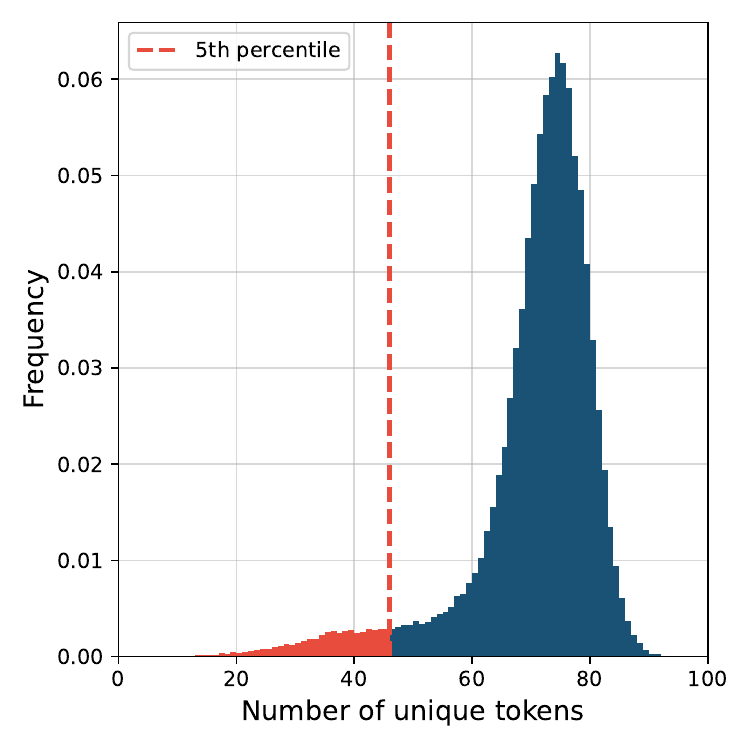}}
    \qquad
    \subfloat[]
    {\includegraphics[width=0.4\textwidth]{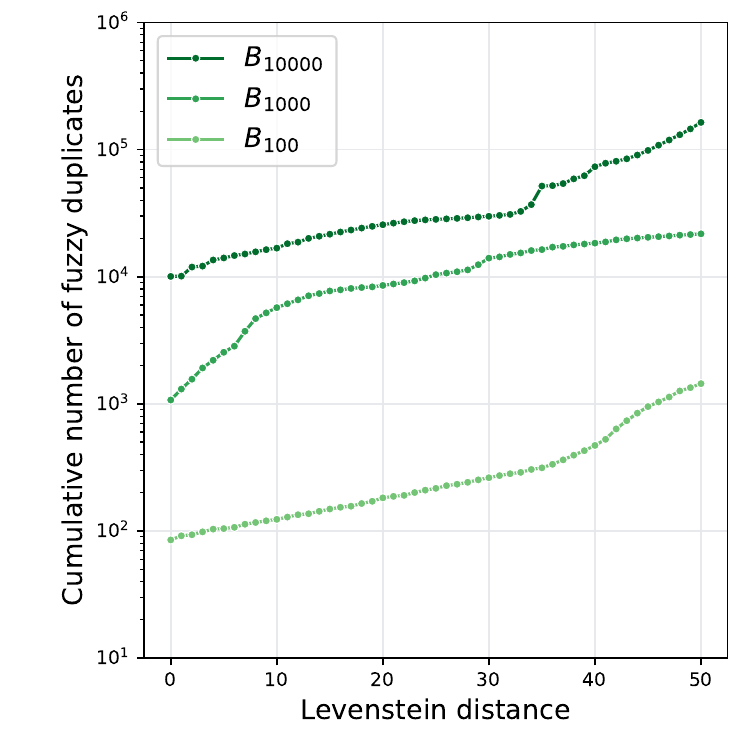}}
    \caption{\textbf{Fuzzy duplicates in SlimPajama.} (a) Distribution of unique tokens across 100-token sequences in SlimPajama. The red line indicates the 5th percentile threshold (46 unique tokens) used to filter out trivial sequences with excessive token repetition. (b) Cumulative number of fuzzy duplicates at increasing Levenshtein distances for sequences with 100, 1,000, and 10,000 exact duplicates in SlimPajama.}
    \label{fig:appendix_slimpajama}
\end{figure*}


We adapt the code from Lee et al.~\cite{lee2022deduplicating} to build a suffix array and find all sequences of 100 tokens repeated more than once in the SlimPajama. We then consider sequences with a certain number of exact repetitions across the dataset: \{100, 1000, 10000\} ($\pm 1\%$). We estimate each of these buckets is large enough to raise concerns if many meaningful fuzzy duplicates are identified. Namely, the dataset has over 50 million sequences repeated at least 100 times, over 700,000 sequences repeated at least 1000 times, and over 30,000 sequences repeated at least 10000 times.

We then randomly pick 100 sequences from each bucket (100, 1000 and 10000 exact repetitions across the dataset) as our target sequences. To focus on meaningful memorization, we apply an additional selection criteria at this stage. Namely, we filter out sequences with too many repetitions of the same token, excluding the sequences where the number of unique tokens is below the 5th percentile across the dataset (46 unique tokens for 100-token sequence, see Fig.~\ref{fig:appendix_slimpajama}(a)). As such, we eliminate trivial sequences in our study, e.g. sequences where one token (often a space) is repeated many times, with only a few non-trivial tokens present. 

Next, we scan the dataset with a moving window of 100 tokens and a 1-token-step, collecting all sequences with the Levenshtein distance to one of the target sequences below a certain threshold (in our case 50). As a computational optimization, we maintain the token overlap counter with the current window for each of the target sequences, allowing us to compute the relatively expensive Levenstein distance in a rare event where the current window has at least 50 common tokens with the target sequence. We then avoid double counting by ensuring that each individual token is contributing to at most one fuzzy duplicate in the final count. To make the computation feasible on a very large dataset such as SlimPajama, we scan the first 5\% of a randomly shuffled dataset and extrapolate the results, assuming fuzzy duplicates are spread mostly uniformly across the shuffled dataset -- which is confirmed by our observations on the positions of the exact duplicates. The scan took 96 hours on 80 CPUs on a server with sufficient memory to fit the whole 5\% portion of the dataset into memory.

\textbf{Additional results.} In the main body of the paper (Figure~\ref{fig:slimpajama}(b)), we provided the number of fuzzy duplicates for target sequences sampled from the bucket of sequences with $1000 \pm 1\%$ exact repetitions. Figure~\ref{fig:appendix_slimpajama}(b) now shows the cumulative number of fuzzy duplicates, averaged across sequences, for the other buckets as well, denoted as $B_{100}$, $B_{1000}$ and $B_{10000}$. Each line starts at the respective number of exact duplicates (which is the selection criteria) and grows as we increase the Levenshtein distance. While the $B_{1000}$ bucket shows the fastest increase, all three buckets reach double the initial number of duplicates ($B_{100}$ by Levenshtein distance 23, $B_{1000}$ by 3 and $B_{10000}$ by 12). Moreover, for higher Levenstein distances, all buckets reach a ten-fold increase compared to the initial number of exact duplicates ($B_{100}$ at Levenstein distance 45, $B_{1000}$ at 24 and $B_{10000}$ at 45). 

In summary, these results show that even in datasets deduplicated on a document level, many exact duplicates still exist. For sequences repeated exactly, we furthermore recover a substantial amount of fuzzy duplicates at values of the Levenshtein distance which correspond to a meaningful contribution to memorization (see Figure~\ref{fig:slimpajama}(a)).

\end{document}